\documentclass{article} 
\usepackage{iclr2027_conference,times}

\usepackage{amsmath,amsfonts,bm}

\def\eqref#1{equation~\ref{#1}}
\def\Eqref#1{Equation~\ref{#1}}

\def\1{\bm{1}}

\DeclareMathAlphabet{\mathsfit}{\encodingdefault}{\sfdefault}{m}{sl}
\SetMathAlphabet{\mathsfit}{bold}{\encodingdefault}{\sfdefault}{bx}{n}

\usepackage{hyperref}
\usepackage{url}
\usepackage{graphicx}
\usepackage{amssymb}

\title{Learning reshapes power-law anisotropy in internal representations}

\author{Asahi Nakamuta \\
  Graduate School of Informatics\\
  Kyoto University\\
  \texttt{nakamuta.asahi.56z@st.kyoto-u.ac.jp} \\
   \And
 Jun-nosuke Teramae \\
  Graduate School of Informatics\\
  Kyoto University\\
  \texttt{teramae@acs.i.kyoto-u.ac.jp} \\
}

\iclrfinalcopy 
\begin{document}

\maketitle

\begin{abstract}
Power-law anisotropy in internal representations has been observed across a wide range of biological and artificial neural systems, from state-of-the-art language models to the mouse cerebral cortex. This anisotropy is a key geometric property of high-dimensional information processing and underlies a variety of theoretical analyses. However, the mechanism by which it emerges from input structure and task-driven learning has remained unclear. Here, we characterize this formation process by exactly solving the learning dynamics of a wide two-layer linear neural network in a teacher--student setting with power-law input and teacher structures. We show that, in the feature-learning regime, the local power-law exponent of the internal-representation spectrum evolves nonmonotonically over the course of training and exhibits up to four distinct asymptotic regimes across modes and training times. By contrast, in the lazy regime, the exponent remains essentially unchanged. We further demonstrate numerically that similar exponent dynamics arise in more realistic nonlinear networks. Together, these results suggest a general mechanism by which the dynamic interaction between input statistics and task structure gives rise to power-law internal representations.
\end{abstract}

\section{Introduction}
Neural networks transform inputs from the external world into high-dimensional internal representations. Understanding the structure of these representations is essential for explaining how networks extract information and form features relevant to a task. One fundamental quantity characterizing this geometry is the eigenspectrum of the covariance matrix, or kernel, of the internal representations.

Recent studies have reported that the eigenvalues of internal-representation covariance matrices decay as a power law with rank in a wide range of neural systems \cite{gauthaman2025universal, morales2024neural, pachitariu2026critical, morales2023quasiuniversal, nakamuta2025self}, including mouse visual cortex \cite{stringer2019high}, computer-vision models \cite{ghosh2022investigating}, and large language models \cite{li2026tracing}. Such power-law anisotropy implies that a large fraction of the variance is concentrated along a small number of directions, while information remains distributed across many higher-order directions. In addition to being observed broadly in prior work, power-law anisotropy has been linked to the efficiency and robustness of information coding \cite{stringer2019high, tatsukawa2025cortical, safavi2024signatures, nassar20201}, generalization performance \cite{ghosh2022investigating}, and to neural scaling laws in training \cite{bordelon2024dynamical, maloney2022solvable, kramp2026dynamics}. Power-law representation spectra may therefore provide an important connection between the microscopic geometry of internal representations and macroscopic performance.

\begin{figure}
\begin{center}
\includegraphics[width=0.8\linewidth]{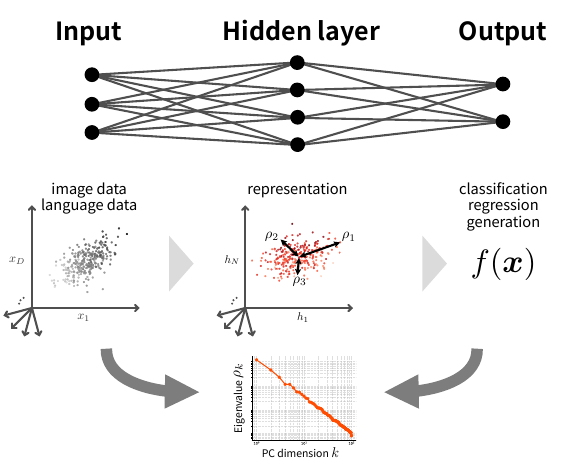}
\end{center}
\caption{Conceptual overview of this study. Previous works have extensively investigated how power-law anisotropy of internal representations affects network outputs and training dynamics, whereas the mechanism by which such representations form has remained insufficiently understood. Here, we clarify how input structure and learning jointly shape these representations.}
\label{fig:overview}
\end{figure}

Most existing theories, however, treat the eigenspectrum of the internal representation as a given quantity that remains fixed during learning. In actual neural networks, the representation itself changes through training. Merely assuming a fixed power-law exponent therefore cannot explain where the power-law spectrum originates or how learning reorganizes it. In particular, because the eigenspectrum of the input covariance of natural data often itself exhibits power-law decay \cite{field1987relations, ruderman1993statistics, stringer2019high, corral2015zipf, font2013scaling}, one must distinguish whether an observed internal representation simply inherits the input structure or is reshaped by task learning (Fig.~\ref{fig:overview}). Although several previous studies have assumed power-law input spectra and derived the dynamics of internal representations \cite{bordelon2025feature, wortsman2025kernel}, to the best of our knowledge, none of them has characterized how the power-law exponent which quantifies the degree of spectral anisotropy evolves during learning. It also remains unclear whether the spectrum retains a single power-law exponent across all modes during learning or instead exhibits multiple local exponents over different mode ranges, and how such changes depend on the learning regime.

\subsection{Our contributions}
To characterize how learning reshapes the eigenspectrum of internal
representation, we study a wide two-layer linear network with aligned power-law input and teacher structures characterized by exponents $\alpha$ and $\beta$, respectively. Our main contributions are as follows:
\begin{enumerate}
    \item \textbf{Exact solution for the time evolution of representation spectra.} We derive an exact analytical expression for the time evolution of the internal-representation eigenspectrum (Result ~\ref{subsec:exact-spectrum}, Eq.~\ref{eq:exact-representation-spectrum}). This expression explicitly decomposes each eigenvalue into an input-derived component present at initialization and a teacher-dependent component formed through learning.
    \item \textbf{Representation spectra exhibit four asymptotic exponent regions during learning.} We analytically characterize the power-law exponent, which quantifies spectral anisotropy (Result ~\ref{subsec:four-exponent-domains} , Eq.~\ref{eq:exact-exponents}, \ref{eq:four-asymptotic-exponents}). In the feature-learning regime, the exponent can evolve nonmonotonically, increasing from $\alpha$ toward $3\alpha+\beta$ before decreasing toward either $\alpha+\beta$ or $\alpha+\beta/2$. By contrast, in the lazy regime, the learning-induced component remains small relative to the initialization component, and the exponent stays close to that of the input.
    \item \textbf{Finite-width nonlinear networks validate the theory.} We validate the theory in finite-width nonlinear networks and observe similar learning-induced changes, confirming that the theoretically predicted spectral transitions are observed beyond the linear model (Result ~\ref{subsec:nonlinear-networks}).
\end{enumerate}

Oveall, our results show that power-law internal representations do not merely reflect the statistical structure of the input data, but are dynamically reshaped by the anisotropy of the teacher signal and by mode-dependent learning timescales. They thus provide an analytical starting point for understanding the power-law internal representations observed across neural networks as a consequence of the interaction between input structure and learning.

\subsection{Related Works}
\addcontentsline{toc}{section}{Related Work}

\paragraph{Power-law spectra of internal representations}

Power-law covariance eigenspectra have been observed in biological neural populations: in mouse V1, the eigenvalues decay approximately as $k^{-1}$, with related scaling reported across species and brain regions \cite{stringer2019high, kong2022increasing, meshulam2019coarse, morales2023quasiuniversal, wang2025geometry}. A recent reanalysis of the mouse V1 data found that a broken power law provides a better fit than a single power law \cite{pospisil2025revisiting}, supporting a local, mode-dependent description of the exponent. Analogous power-law feature-covariance spectra and their evolution during training have been empirically observed in vision models \cite{agrawal2022alpha, ghosh2022investigating} and large language models \cite{li2026tracing}.

\paragraph{Input structure and spectral theories of fixed representations}

Power-law decay of covariance spectra has long been reported for natural images as well \cite{field1987relations, ruderman1993statistics}. However, the persistence of a power-law spectrum in V1 responses even when the input images are spatially whitened indicates that internal representations cannot be explained solely by direct inheritance of the input covariance \cite{stringer2019high}. In random-feature models, the input power-law exponent has been shown to carry over to the feature-covariance spectrum up to logarithmic corrections \cite{paquette2026power}. Theories based on fixed kernels or fixed features have used eigenspectra and target alignment to derive learning order, generalization curves, and neural scaling laws \cite{kramp2026dynamics, bordelon2024dynamical, bordelon2025feature, maloney2022solvable}. Beyond random-feature models, however, it remains unknown how training deforms an inherited input exponent when the internal kernel evolves over time.

\paragraph{Learning dynamics of deep linear networks}

Although deep linear networks implement linear input--output maps, the gradient dynamics of their factorized weights are nonlinear, and these networks have therefore been studied as analytically tractable feature-learning models. Exact learning curves for aligned singular modes have been derived, revealing plateaus and stage-like learning \cite{saxe2013exact}. The exact temporal evolution of the hidden representational similarity matrix and the neural tangent kernel was subsequently derived as well \cite{braun2022exact, domine2025lazy}. Building on these results, we derive the hidden-representation covariance in the population limit for power-law input and teacher structures that are simultaneously diagonalizable, with particular emphasis on the power-law decay of its eigenspectrum.

\section{Setting}

We consider a mean-field parameterization of a two-layer linear neural network in a teacher-student setting with input dimension \(D\), hidden width \(N\), and output dimension \(P\):
\begin{align}
    \boldsymbol{f}(\boldsymbol{x},t)
    &=
    \frac{1}{\gamma_0 N}
    \sum_{i=1}^{N}
    \boldsymbol{a}_i(t)\,
    \boldsymbol{w}_i(t)^{\top}\boldsymbol{x},
    \label{eq:student-network}
    \\
    \boldsymbol{y}(\boldsymbol{x})
    &=
    \boldsymbol{\Theta}^{\top}\boldsymbol{x}.
    \label{eq:teacher-map}
\end{align}
Here, \(\boldsymbol{y}(\boldsymbol{x})\) denotes teacher output, and \(\boldsymbol{\Theta}\) is teacher matrix. The dimensions of the variables are
\(\boldsymbol{x},\boldsymbol{w}_i\in\mathbb{R}^{D}\),
\(\boldsymbol{a}_i,\boldsymbol{f},\boldsymbol{y}\in\mathbb{R}^{P}\), and
\(\boldsymbol{\Theta}\in\mathbb{R}^{D\times P}\).
We assume
\begin{equation}
    P=D\ll N,
    \qquad
    N\to\infty.
\end{equation}
Thus, the input and output dimensions take the same finite value, while the hidden-layer width is taken to infinity. Although conventional regression and classification problems typically have $P\ll D$, the choice $P\ge D$ is natural for architectures such as large language models, diffusion models, and autoencoders. As will become clear below, $P\ge D$ is also required for the spectrum to evolve over a broad range of modes. We therefore focus on the simplest such setting, $P=D$. Following previous work \cite{bordelon2022self, bordelon2024depthwise, bordelon2024infinite}, the parameter $\gamma_0$ controls the learning regime. Although this deep linear model can represent only linear input--output maps, its learning dynamics are nonlinear and exhibit a variety of rich phenomena, making it an important object of study in machine-learning theory.

We assume that the input data $\boldsymbol{x}$ and the teacher matrix $\boldsymbol{\Theta}$ have the following aligned structure: the input covariance matrix and the teacher Gram matrix are diagonal in the same orthonormal basis, and their eigenvalues decay as power laws with exponents $\alpha$ and $\beta$, respectively:
\begin{align}
    \boldsymbol{x}
    &\sim
    \mathcal{N}\!\left(\boldsymbol{0},\boldsymbol{\Lambda}\right),
    \label{eq:input-distribution}
    \\
    \boldsymbol{\Lambda}
    &=
    \operatorname{diag}\!\left(\lambda_1,\ldots,\lambda_D\right),
    \label{eq:input-covariance}
    \\
    \lambda_k
    &\sim
    k^{-\alpha},
    \label{eq:input-power-law}
    \\
    \boldsymbol{\Theta}\boldsymbol{\Theta}^{\top}
    &=
    \operatorname{diag}\!\left(\mu_1,\ldots,\mu_D\right),
    \label{eq:teacher-gram}
    \\
    \mu_k
    &\sim
    k^{-\beta}.
    \label{eq:teacher-power-law}
\end{align}

The weights \(\boldsymbol{w}_i(t)\) and \(\boldsymbol{a}_i(t)\) are initialized independently from Gaussian distributions with variances \(\sigma_w^2\) and \(\sigma_a^2\), respectively, and are trained by gradient flow on the mean-squared-error loss with learning rate \(\nu\). Although the temporal evolution of the parameters and loss is known from previous work \cite{saxe2013exact}, neither the power-law anisotropy of the internal representation nor the evolution of its exponent has been studied. We therefore analytically derive the temporal evolution of the eigenvalues of the internal-representation covariance in the infinite-data limit and characterize the corresponding power-law exponents.

\section{Results}
\label{sec:results}

\begin{figure}
\begin{center}
\includegraphics[width=0.9\linewidth]{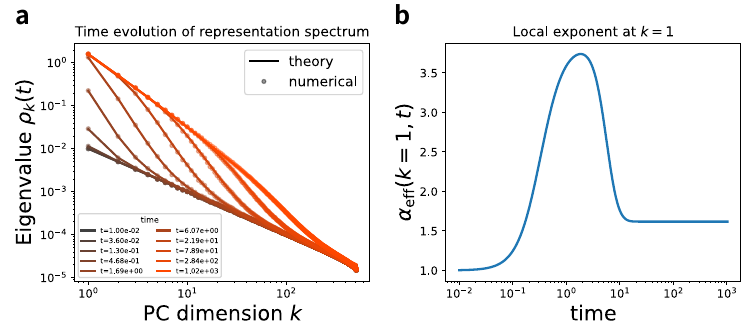}
\end{center}
  \caption{Evolution of the internal representations of a two-layer linear model whose
power-law exponent changes during training.
\emph{(a)} Eigenvalue spectra of the internal representation at different
training times. Solid lines denote the analytical solution, markers denote
the numerical solution, and color indicates training time.
\emph{(b)} Local power-law exponent at each time, evaluated at \(k=1\);
equivalently, this is the slope of the spectrum on logarithmic axes.}
  \label{fig:representation-spectra}
\end{figure}

\subsection{Exact solution for the time evolution of representation spectra}
\label{subsec:exact-spectrum}

We first derive the time evolution of the internal-representation eigenvalue
spectrum in the two-layer linear network. The internal-representation kernel
is defined as
\begin{equation}
    \Phi_t(\boldsymbol{x},\boldsymbol{x}')
    =
    \boldsymbol{x}^{\top}
    \boldsymbol{G}^{w}(t)
    \boldsymbol{x}',
    \label{eq:representation-kernel}
\end{equation}
where
\begin{equation}
    \boldsymbol{G}^{w}(t)
    =
    \frac{1}{N}
    \sum_{i=1}^{N}
    \boldsymbol{w}_i(t)\boldsymbol{w}_i(t)^{\top}
    \label{eq:first-layer-gram}
\end{equation}
is the Gram matrix of the first-layer weights. Although our object of
interest is the spectrum of \(\Phi_t\), in the population limit its nonzero
eigenvalues coincide with those of
\begin{equation}
    \boldsymbol{M}(t)
    =
    \boldsymbol{\Lambda}^{1/2}
    \boldsymbol{G}^{w}(t)
    \boldsymbol{\Lambda}^{1/2}
    \label{eq:population-spectrum-matrix}
\end{equation}
(Appendix~\ref{app:operator}). We therefore analyze the eigenvalue spectrum
\(\rho_k(t)\) of \(\boldsymbol{M}(t)\).

In our setting, because the input covariance and the teacher Gram matrix are diagonalized by the same basis, the eigenvalues $\rho_k$ evolve independently mode by mode and are given exactly by
\begin{equation}
\begin{aligned}
    \rho_k(t)
    &=
    \underbrace{\sigma_w^2\lambda_k}_{
        \substack{\text{input-derived}\\\text{initialization component}}
    }
    +
    \underbrace{
        \sigma_w^2\lambda_kR(u_k(t))
    }_{
        \substack{\text{teacher-derived component}\\
        \text{formed through learning}}
    },
    \\
    R(u_k(t))
    &=
    \frac{S}{2\sigma_w^2}
    \left[
        -1+
        \sqrt{1+4u_k(t)^2}
    \right],
    \\
    u_k(t)
    &=
    \frac{v_k(t)}{1-v_k(t)^2},
    \\
    v_k(t)
    &=
    \frac{r_k^+\bigl(E_k(t)-1\bigr)}
    {(r_k^+)^2+E_k(t)},
    \\
    E_k(t)
    &=
    \exp\left(
        \frac{
            \lambda_k
            \sqrt{S^2+4\gamma_0^2\theta_k^2}
        }{\gamma_0^2}
        t
    \right),
    \\
    r_k^+
    &=
    \frac{
        -S+\sqrt{S^2+4\gamma_0^2\theta_k^2}
    }{2\gamma_0\theta_k},
\end{aligned}
\label{eq:exact-representation-spectrum}
\end{equation}
where \(S=\sigma_w^2+\sigma_a^2\) and \(\theta_k=\sqrt{\mu_k}\)
(Appendix~\ref{app:both-layers-solution}). This derivation uses the
wide-network limit \(P=D\ll N\), with \(N\to\infty\). Throughout the evolution described above, $\rho_k(t)$ remains monotonically decreasing in $k$; hence, the relative ordering of the modes is preserved at all times (Appendix~\ref{app:ordering}). The input-mode index $k$ can therefore be identified directly with the rank index of the spectrum.

\Eqref{eq:exact-representation-spectrum} shows that the spectrum
consists of two components. The first is inherited from the input data and
is already present at initialization, whereas the second is induced by the
teacher signal through learning. Thus, the anisotropy of the internal
representation evolves not only according to the anisotropy of the input
distribution but also according to the anisotropy of the teacher signal and
the learning dynamics.

The eigenspectrum dynamics predicted by the analytical solution above closely match those obtained numerically for finite-width networks
(Fig.~\ref{fig:representation-spectra}(a), Appendix~\ref{app:numerical-settings}). At initialization, the spectrum
\(\rho_k(0)\) has the same power-law exponent \(\alpha\) as the input
variance \(\lambda_k\). As training proceeds, the modes acquire the
teacher-derived component in order from low to high mode index. The evolution of the spectrum demonstrates that the local effective
power-law exponent $\alpha_{\mathrm{eff}}(k,t)$ can change during training
(Fig.~\ref{fig:representation-spectra}(b). We define this exponent
formally and characterize its temporal evolution in the next subsection.

\begin{figure}
\begin{center}
\includegraphics[width=1.0\linewidth]{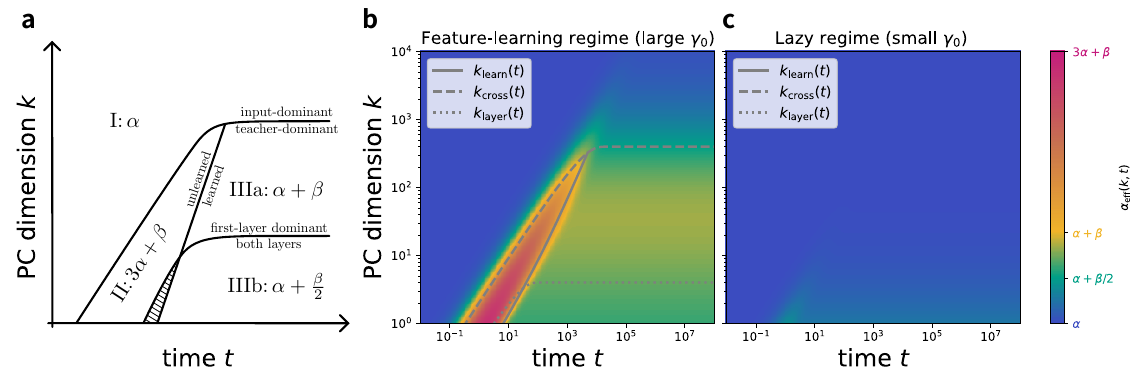}
\end{center}
  \caption{Evolution of the local power-law exponent \(\alpha_\mathrm{eff}\), defined as the negative slope
of the eigenspectrum on log--log axes.
In all panels, the horizontal axis represents training time and the
vertical axis represents PC dimension (mode rank).
In the heatmaps, color indicates the local exponent \(\alpha_\mathrm{eff}(k,t)\) around each mode \(k\) at each training time \(t\).
\emph{(a)} Schematic of the plateaus and their boundaries in the local
exponent around each mode during training. The hatched regions exhibit
decay that is not described by a single power law.
\emph{(b)} In the feature-learning regime, the internal-representation
spectrum exhibits as many as four asymptotic exponents.
\emph{(c)} In the lazy regime, the exponent of the internal representation
changes very little.}
  \label{fig:exponent-regions}
\end{figure}

\subsection{Representation spectra exhibit four asymptotic exponent domains during learning}
\label{subsec:four-exponent-domains}

We next investigate how the power-law exponent of the eigenvalue spectrum,
which characterizes the anisotropy of the internal representation, changes
during learning. After taking the infinite-width limit \(N\to\infty\), we consider
sufficiently large input and output dimensions \(D\) and \(P\), and assume
that, over the mode range \(1\leq k\leq D\), the input variance and the
mode strength of the teacher matrix obey
\begin{equation}
    \lambda_k=\lambda_0 k^{-\alpha},
    \qquad
    \mu_k=\theta_k^2=\mu_0 k^{-\beta}.
    \label{eq:mode-power-laws}
\end{equation}
We define the local effective power-law exponent of the representation
spectrum, which depends on time \(t\) and mode \(k\), by
\begin{equation}
    \alpha_{\mathrm{eff}}(k,t)
    :=
    -\frac{\partial\log\rho_k(t)}{\partial\log k}.
    \label{eq:effective-exponent}
\end{equation}
Here, the derivative is defined after continuously interpolating the
discrete sequence of modes.

For the setting considered here, the local exponent $\alpha_{\mathrm{eff}}(k,t)$ takes different values in distinct domains of the \((k,t)\) plane (Fig.~\ref{fig:exponent-regions}). We therefore first determine the boundaries separating these domains and then derive the asymptotic value of the local power-law exponent within each domain.

Substituting the analytical expression for the eigenspectrum derived in Result ~\ref{subsec:exact-spectrum} (Eq. 12) into the definition of the power-law exponent yields the following exact expression (Appendix~\ref{app:boundaries}):

\begin{equation}
\left\{
\begin{aligned}
\alpha_{\mathrm{eff}}(k,t)
&= \alpha 
+ \frac{R(u_k(t))}{1+R(u_k(t))}
\left(
1+\frac{1}{g(u_k(t))}
\right)
p(k,t)\\
p(k,t) &= \frac{(u_{k,\infty}-u_k(t))g(u_k(t))}{u_k(t)}
\int_0^{u_k(t)}
\left[
\frac{\alpha}
{u_{k,\infty}-z}
+ \frac{u_{k,\infty}\beta/2}
{(u_{k,\infty}-z)^2}
\right]
\frac{dz}{g(z)}
\end{aligned}
\right.
\label{eq:exact-exponents}
\end{equation}

where $g(x) := \sqrt{1+4x^2}, u_{k,\infty} := \gamma_0 \theta_k/S$.

The form of the exact solution above indicates that the main changes in the power-law exponent occur when \(R(u_k(t)) = 1\), \(u_k(t) \rightarrow u_{k,\infty}\), and \(u_k(t) = 1\). Accordingly, we define the boundaries between the asymptotic domains as follows (Appendix~\ref{app:boundaries}):
\begin{itemize}
    \item \(k_{\mathrm{cross}}(t)\) is defined by
    \(R(u_k(t))=1\), or equivalently $u_k(t) = u_{\mathrm{cross}}:= \sqrt{\sigma_w^2(S+\sigma_w^2)}/S$, separating the input-dominated region from the teacher-dominated region;
    \item \(k_{\mathrm{learn}}(t)\) is defined by \(u_k(t)=0.9u_{k,\infty}\), separating the unlearned region from the learned region; and
    \item \(k_{\mathrm{layer}}(t)\) is defined by
    \(u_k(t)=1\), separating the region in which
only the first layer learns from that in which both layers learn.
\end{itemize}

Near these boundaries, where $u_k(t)$ is close to the corresponding
threshold, the spectrum is generally not described by a single power
law. Sufficiently far from the boundaries, however, the local exponent
approaches a constant asymptotic value. The regions are characterized
as follows (Fig.~\ref{fig:exponent-regions} and Appendix~\ref{app:boundaries}).

\begin{itemize}
    \item \textbf{Region I: Initialization-dominated.}
    Above $k_{\mathrm{cross}}(t)$, the learning-induced change in the
    internal representation is smaller than the initialization
    component. The exponent approaches $\alpha$ in this region. This domain contains not only unlearned modes but also modes for which the teacher signal is so weak that the initialization component remains dominant even after learning. In particular, all sufficiently large k belong to this domain. Conse-quently, changes in the local power-law exponent are observed only over a finite range of modes.

    \item \textbf{Region II: Transient steepening during early learning.}
    Below $k_{\mathrm{cross}}(t)$ and above $k_{\mathrm{learn}}(t)$, with $u_k(t) \ll 1$, modes have only recently begun to learn, yet the learning-induced component already exceeds the initialization component. The exponent rises toward $3\alpha+\beta$. By contrast, in a domain where \(u_k(t)\gg 1\), the eigenvalue spectrum \(\rho_k(t)\) then has no single power-law exponent.

    \item \textbf{Region IIIa: Learned modes with first-layer-dominated learning.}
    Below both $k_{\mathrm{cross}}(t)$ and $k_{\mathrm{learn}}(t)$,
    but above $k_{\mathrm{layer}}(t)$, the exponent approaches
    $\alpha+\beta$. A necessary condition for this asymptotic region
    to exist is that the initialization scale of the second layer
    be much larger than that of the first, $\sigma_a \gg \sigma_w$
    (Appendix~\ref{app:region-existence-conditions}). The analytical solution in Appendix~\ref{app:both-layers-solution} shows that
    learning in this region substantially changes the first layer
    relative to its initialization. Region IIIa therefore corresponds
    to learning primarily in the first layer.

    \item \textbf{Region IIIb: Learned modes with learning in both layers.}
    Within the learned, teacher-dominated region, modes below
    $k_{\mathrm{layer}}(t)$ approach an exponent of $\alpha+\beta/2$.
    In contrast to Region IIIa, both layers learn in this region.
\end{itemize}

Collecting the results above, the internal-representation spectrum in the
\((k,t)\) plane exhibits as many as four asymptotic exponents in bulk
domains sufficiently far from the boundaries
(Fig.~\ref{fig:exponent-regions}(a, b)):
\begin{equation}
    \alpha_{\mathrm{eff}}(k,t)
    \simeq
    \begin{cases}
        \alpha,
        &
        u_k(t)\ll u_\mathrm{cross},
        \\[4pt]
        3\alpha+\beta,
        &
        u_k(t)\gg u_\mathrm{cross},\quad
        u_k(t)\ll u_{k,\infty},\quad
        u_k(t)\ll1,
        \\[4pt]
        \alpha+\beta,
        &
        u_k(t)\gg u_\mathrm{cross},\quad
        u_k(t)\simeq u_{k,\infty},\quad
        u_k(t)\ll1,
        \\[4pt]
        \alpha+\dfrac{\beta}{2},
        &
        u_k(t)\gg u_\mathrm{cross},\quad
        u_k(t)\simeq u_{k,\infty},\quad
        u_k(t)\gg1.
    \end{cases}
    \label{eq:four-asymptotic-exponents}
\end{equation}

When a fixed mode \(k\) is followed over time, its location relative to
\(k_{\mathrm{layer}}\) determines which of the following sequences it
exhibits:
\begin{equation}
    \alpha
    \longrightarrow
    3\alpha+\beta
    \longrightarrow
    \begin{cases}
        \alpha+\beta,\\
        \alpha+\dfrac{\beta}{2}.
    \end{cases}
    \label{eq:fixed-mode-exponent-sequence}
\end{equation}
Whether all four regions exist depends on the hyperparameters. In particular, in the feature-learning regime characterized by 
$\sigma_a \gg \sigma_w
\text{ and }
\gamma_0 \gg (\sigma_a^2+\sigma_w^2)/\theta_0
$
and for sufficiently larger \(D\), 4
all four regions can coexist (see Appendix~\ref{app:boundaries}).
Consequently, the local effective exponent
first steepens transiently and then decreases. This nonmonotonic change
reflects the mode-dependent learning rates set by the input eigenvalues and the process by which the anisotropy of the teacher signal is incorporated into the internal representation.

By contrast, in the lazy regime satisfying
$
\gamma_0 \ll \sqrt{\sigma_w^2(2\sigma_w^2+\sigma_a^2)}/\theta_0
$,
the learned component does not exceed the initialization component even for
the strongest teacher mode. Hence \(R(u_k(t))\ll1\) is maintained for every
mode, and
\begin{equation}
    \rho_k(t)\simeq\sigma_w^2\lambda_k,
    \qquad
    \alpha_{\mathrm{eff}}(k,t)\simeq\alpha
    \label{eq:lazy-regime-spectrum}
\end{equation}
(Fig.~\ref{fig:exponent-regions}(c)). The nonmonotonic exponent evolution
and the multiple bulk exponents derived above are therefore specific to the
feature-learning regime, in which the internal representation changes
substantially from its initial value.

\subsection{Similar exponent dynamics appear in nonlinear networks}
\label{subsec:nonlinear-networks}

Finally, to test whether the exponent changes found in the solvable model
also arise in more general models, we perform a numerical study using
nonlinear networks. We train a network with a nonlinear activation function
in a teacher--student setting with the same type of power-law input
variance and teacher signal, and measure the temporal evolution of the
internal-representation spectrum (Appendix~\ref{app:numerical-settings}).

\begin{figure}
\begin{center}
\includegraphics[width=0.9\linewidth]{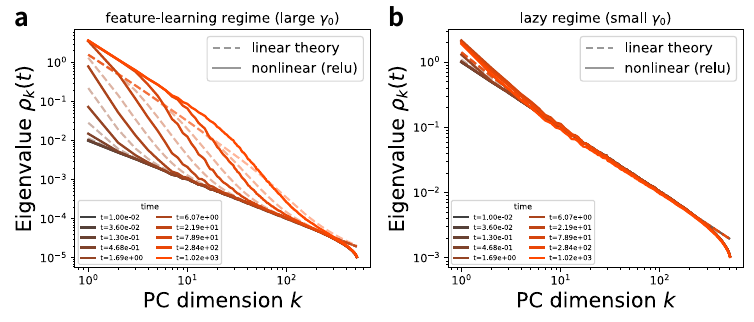}
\end{center}
  \caption{Numerical validation using a nonlinear student network. As training
progresses, the slope of the internal-representation spectrum changes in
the feature-learning regime but remains close to its initial value in the
lazy regime.}
  \label{fig:nonlinear-validation}
\end{figure}

The slope of the internal-representation spectrum also changes over the
course of learning in the nonlinear network
(Fig.~\ref{fig:nonlinear-validation}). In particular, in the
feature-learning regime, the spectrum changes successively from low to high
mode index, and its exponent evolves from \(\alpha\) to
\(\alpha+\beta\) over a finite range of modes. In the lazy regime, by
contrast, it remains at \(\alpha\). These results agree quantitatively with
the predictions of the linear theory.

This result suggests that the mechanism analyzed exactly in the two-layer
linear model is not peculiar to linear networks, but may capture a basic
mechanism of internal-representation formation in a broader class of
neural networks.

\section{Conclusion}
\label{sec:conclusion}

We analyzed how a wide, two-layer linear neural network reshapes the
spectrum of its internal representations during learning in a teacher--student setting where
the input data and the teacher map have aligned power-law structure. We
derived the exact temporal evolution of the eigenvalues of the hidden
representation kernel in the population limit and showed that the
internal-representation spectrum decomposes into a baseline component
inherited from the input covariance at initialization and an increment
formed through training.

Using this decomposition, we showed that, in the feature-learning regime
and over finite mode ranges sufficiently far from the boundaries, the local
effective power-law exponent can take four asymptotic values:
\begin{equation}
    \alpha,
    \qquad
    3\alpha+\beta,
    \qquad
    \alpha+\beta,
    \qquad
    \alpha+\frac{\beta}{2}.
\end{equation}
These exponents correspond, respectively, to dominance by the initialization
component, mode-dependent representation changes at early learning times,
post-learning deformation driven primarily by the first layer, and
deformation distributed across both layers. In particular, at early
learning times, the mode-dependent learning rates set by the input
eigenvalues transiently steepen the spectrum. The spectrum then transitions
to an exponent determined by the strength of the teacher map and the
allocation of learning across layers. In the lazy regime, by contrast, the
training-induced component never exceeds the initialization component, and
the power-law exponent of the internal representation remains nearly
unchanged from the input exponent \(\alpha\).

Numerical experiments with nonlinear networks further showed that the main
exponent transitions predicted by the theory, as well as the difference
between the feature-learning and lazy regimes, persist beyond the linear
model. This result suggests that the interaction among input anisotropy, the
modal structure of the teacher signal, and mode-dependent learning rates may
also govern internal-representation formation in more general networks.

Our analysis relies on simultaneous diagonalizability of the input
covariance and the teacher matrix, infinite width, and population-limit
gradient flow. Future work should extend the analysis to misaligned input
and teacher structure, finite-width and finite-sample effects, deeper
networks, and general nonlinear activation functions. It will also be
important to determine the extent to which the predicted exponent
transitions and boundaries can be observed in real data and large-scale
models.

Taken together, these results show that power-law internal representations
should not be regarded as possessing a single fixed exponent. Rather, their
exponents are reshaped over time and across modes through the interaction
between input statistics and task learning. This work provides a foundation
for analytically understanding how power-law representations are inherited
and transformed in neural networks.

\subsection*{AI use statement}
In this work, we used generative AI tools to help develop theoretical models or conceptual frameworks, formulate mathematical claims, provide critical ingredients for proving mathematical claims, assist in the writing of proofs, propose or refine hypotheses, design or provide feedback on research  methodology or experiments, implement methods, and assist with translation.
We have not used generative AI tools to generate synthetic datasets, clean and reformat dataset, support qualitative and thematic data analysis, or interpret results.
Additionally, we used generative AI tools to draft parts of a research paper, summarize or analyse existing literature, edit a research paper to improve readability, and identify relevant literature. We have reviewed all AI-assisted work. We manually checked the manuscript text, analytical calculations, numerical computations, and cited references. We take responsibility for the final content of this work,
including text, claims or artifacts produced with the aid of generative AI.

\subsection*{Reproducibility statement}
Mathematical details are provided in the appendix section.
The code used to generate the figures is provided in the supplementary material.


\bibliography{iclr2027_conference}
\bibliographystyle{iclr2027_conference}

\newpage

\appendix
\section*{Appendix}
\addcontentsline{toc}{section}{Appendix}
Throughout the appendices, we define the following three Gram matrices:

\begin{align}
\boldsymbol{G}^{w}(t)
&:=
\frac{1}{N}
\sum_{i=1}^{N}
\boldsymbol{w}_i(t)
\boldsymbol{w}_i(t)^{\top},
\\
\boldsymbol{G}^{a}(t)
&:=
\frac{1}{N}
\sum_{i=1}^{N}
\boldsymbol{a}_i(t)
\boldsymbol{a}_i(t)^{\top},
\\
\boldsymbol{G}^{wa}(t)
&:=
\frac{1}{N}
\sum_{i=1}^{N}
\boldsymbol{w}_i(t)
\boldsymbol{a}_i(t)^{\top}.
\end{align}

\section{The anisotropy of the internal representation corresponds to the eigenvalues of \texorpdfstring{$\boldsymbol{M}(t)$}{M(t)}}
\label{app:operator}

The kernel of the internal representation can be written as
\begin{equation}
\Phi_t(\boldsymbol{x},\boldsymbol{x}')
=
\boldsymbol{x}^{\top}
\boldsymbol{G}^{w}(t)
\boldsymbol{x}'.
\end{equation}
We wish to determine the eigenvalues of this kernel. In particular, in the population limit, where the number of samples is infinity, empirical averages can be replaced by expectations over the data distribution. The kernel eigenvalue problem then becomes equivalent to that of the following integral operator:
\begin{equation}
(T_{\Phi}u)(\boldsymbol{x})
=
\mathbb{E}_{\boldsymbol{x}'}
\left[
\boldsymbol{x}^{\top}
\boldsymbol{G}^{w}(t)
\boldsymbol{x}'
u(\boldsymbol{x}')
\right]
=
\rho u(\boldsymbol{x}).
\end{equation}
Because the left-hand side is linear in \(\boldsymbol{x}\), every eigenfunction \(u(\boldsymbol{x})\) associated with a nonzero eigenvalue is also linear in \(\boldsymbol{x}\). We use this fact to derive a matrix \(\boldsymbol{M}(t)\) whose eigenvalues coincide with the nonzero eigenvalues of this operator.

Let
\begin{equation}
\psi_k(\boldsymbol{x})
=
\frac{x_k}{\sqrt{\lambda_k}}
\end{equation}
be an orthonormal basis for the subspace of linear functions. These functions are indeed orthonormal under the expectation inner product on the function space containing \(u(\boldsymbol{x})\). The action of \(T_{\Phi}\) on this basis is
\begin{align}
(T_{\Phi}\psi_l)(\boldsymbol{x})
&=
\mathbb{E}_{\boldsymbol{x}'}
\left[
\boldsymbol{x}^{\top}
\boldsymbol{G}^{w}
\boldsymbol{x}'
\frac{x_l'}{\sqrt{\lambda_l}}
\right]
\\
&=
\boldsymbol{x}^{\top}
\boldsymbol{G}^{w}
\sqrt{\lambda_l}\boldsymbol{e}_l
\\
&=
\sum_k x_k G^{w}_{kl}\sqrt{\lambda_l}
\\
&=
\sum_k
\sqrt{\lambda_k}
G^{w}_{kl}
\sqrt{\lambda_l}
\psi_k(\boldsymbol{x}).
\end{align}
Thus, defining
\begin{equation}
\boldsymbol{M}(t)
:=
\boldsymbol{\Lambda}^{1/2}
\boldsymbol{G}^{w}(t)
\boldsymbol{\Lambda}^{1/2},
\end{equation}
we obtain the matrix representation
\begin{equation}
T_{\Phi}\psi_l
=
\sum_k M_{kl}\psi_k.
\end{equation}

Using the fact that the eigenfunction $u(\boldsymbol{x})$ is linear, we can write $u(\boldsymbol{x})=\sum_l c_l\psi_l(\boldsymbol{x})$, and hence
\begin{align}
T_{\Phi}u
&=
T_{\Phi}
\left(
\sum_l c_l\psi_l
\right)
\\
&=
\sum_l c_lT_{\Phi}\psi_l
\\
&=
\sum_l c_l\sum_kM_{kl}\psi_k
\\
&=
\sum_k
\left(
\sum_lM_{kl}c_l
\right)
\psi_k.
\end{align}
Therefore, the eigenvalue problem for \(T_{\Phi}\) becomes
\begin{equation}
\sum_k
\left(
\sum_lM_{kl}c_l
\right)
\psi_k
=
\rho\sum_kc_k\psi_k.
\end{equation}
Since the \(\psi_k\) are linearly independent, uniqueness of the basis expansion gives
\begin{align}
\sum_lM_{kl}c_l
&=
\rho c_k,
\\
\therefore\qquad
\boldsymbol{M}\boldsymbol{c}
&=
\rho\boldsymbol{c}.
\end{align}
This is precisely the eigenvalue problem for \(\boldsymbol{M}\). Hence, the nonzero eigenvalues of \(T_{\Phi}\) coincide with those of \(\boldsymbol{M}(t)\). Consequently, the anisotropy of the internal representation can be characterized by the eigenvalues of
\begin{equation}
\boldsymbol{M}(t)
=
\boldsymbol{\Lambda}^{1/2}
\boldsymbol{G}^{w}(t)
\boldsymbol{\Lambda}^{1/2}.
\end{equation}

\section{Warm-up: Derivation of the eigenvalues when the second layer is fixed}
\label{app:frozen-second-layer}

As a warm-up, we solve the time evolution of the first-layer weights \(\boldsymbol{w}_i(t)\) when the second-layer weights \(\boldsymbol{a}_i\) are fixed and only the first layer is trained. This yields an analytical solution for \(\boldsymbol{M}(t)\). As shown below, because \(\boldsymbol{\Theta}\boldsymbol{\Theta}^{\top}\) is assumed to be diagonal, \(\boldsymbol{M}(t)\) remains diagonal through training, and the dynamics of its eigenvalues \(\rho_k\) decouple across \(k\).

The student network can be written as
\begin{equation}
\boldsymbol{f}(\boldsymbol{x},t)
=
\frac{1}{\gamma_0}
\boldsymbol{G}^{wa}(t)^{\top}
\boldsymbol{x}.
\end{equation}
In the population limit, define
\begin{equation}
\boldsymbol{R}(t)
:=
\boldsymbol{\Theta}
-
\frac{1}{\gamma_0}
\boldsymbol{G}^{wa}(t).
\end{equation}
The loss is then
\begin{align}
L(t)
&=
\frac{1}{2}
\mathbb{E}_{\boldsymbol{x}}
\left[
\left\lVert
\boldsymbol{f}(\boldsymbol{x},t)
-
\boldsymbol{y}(\boldsymbol{x})
\right\rVert^2
\right]
\\
&=
\frac{1}{2}
\mathbb{E}_{\boldsymbol{x}}
\left[
\sum_p
\left(
\sum_kR_{kp}(t)x_k
\right)^2
\right]
\\
&=
\frac{1}{2}
\sum_p\sum_k
R_{kp}(t)\lambda_kR_{kp}(t)
\\
&=
\frac{1}{2}
\operatorname{Tr}
\left[
\boldsymbol{R}(t)^{\top}
\boldsymbol{\Lambda}
\boldsymbol{R}(t)
\right].
\end{align}
We now determine the time evolution of \(\boldsymbol{M}(t)\) under gradient-flow training in the infinite-data limit.

The first-layer weights obey
\begin{align}
\dot{w}_{ik}(t)
&=
-\nu
\frac{\partial L}{\partial w_{ik}}
\\
&=
-\nu
\sum_{p=1}^{P}
\frac{\partial L}{\partial G^{wa}_{kp}}
\frac{\partial G^{wa}_{kp}}{\partial w_{ik}}
\\
&=
\frac{\nu}{\gamma_0}
\sum_{p=1}^{P}
\lambda_kR_{kp}(t)\frac{\partial G^{wa}_{kp}}{\partial w_{ik}}
\\
&=
\frac{\nu}{\gamma_0N}
\sum_{p=1}^{P}
\lambda_kR_{kp}(t)a_{ip}
\\
&=
\frac{1}{\gamma_0}
\sum_{p=1}^{P}
\lambda_kR_{kp}(t)a_{ip},
\\
\therefore\qquad
\dot{\boldsymbol{w}}_i(t)
&=
\frac{1}{\gamma_0}
\boldsymbol{\Lambda}
\boldsymbol{R}(t)
\boldsymbol{a}_i,
\end{align}
where we set the learning rate to \(\nu=N\) so that the parameter dynamics remain \(O(1)\) as \(N\to\infty\).

Combining this equation with the definition of \(\boldsymbol{G}^{wa}\) gives
\begin{align}
\dot{\boldsymbol{G}}^{wa}(t)
&=
\frac{1}{\gamma_0N}
\boldsymbol{\Lambda}
\boldsymbol{R}(t)
\sum_i
\boldsymbol{a}_i\boldsymbol{a}_i^{\top}
\\
&=
\frac{\sigma_a^2}{\gamma_0}
\boldsymbol{\Lambda}
\left(
\boldsymbol{\Theta}
-
\frac{1}{\gamma_0}
\boldsymbol{G}^{wa}(t)
\right).
\end{align}
Here, using \(N\gg D\), we set
\begin{equation}
\frac{1}{N}
\sum_i
\boldsymbol{a}_i\boldsymbol{a}_i^{\top}
=
\sigma_a^2\boldsymbol{I}.
\end{equation}
Solving the differential equation for \(\boldsymbol{G}^{wa}\) yields
\begin{equation}
G^{wa}_{kp}(t)
=
\gamma_0\Theta_{kp}
+
\left[
G^{wa}_{kp}(0)
-
\gamma_0\Theta_{kp}
\right]
\exp
\left(
-\frac{\sigma_a^2}{\gamma_0^2}
\lambda_kt
\right).
\end{equation}

\begin{figure}
  \centering
  \includegraphics[width=0.4\linewidth]{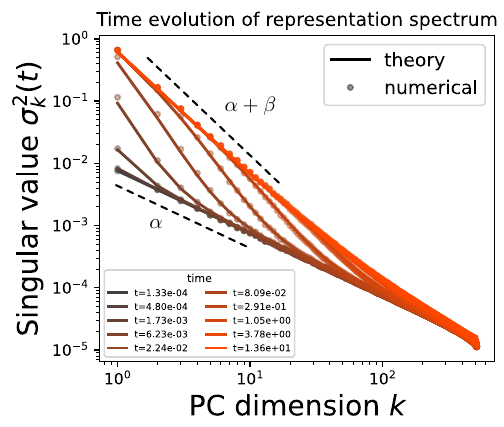}
  \caption{Eigenvalue dynamics for the model with a fixed second layer.}
  \label{fig:frozen-spectrum}
\end{figure}

We next obtain the time evolution of \(\boldsymbol{M}(t)\). From
\begin{equation}
\boldsymbol{G}^{w}(t)
:=
\frac{1}{N}
\sum_{i=1}^{N}
\boldsymbol{w}_i(t)\boldsymbol{w}_i(t)^{\top},
\end{equation}
we have
\begin{align}
\dot{\boldsymbol{G}}^{w}(t)
&=
\frac{1}{N}
\sum_i
\left[
\boldsymbol{w}_i(t)
\dot{\boldsymbol{w}}_i(t)^{\top}
+
\dot{\boldsymbol{w}}_i(t)
\boldsymbol{w}_i(t)^{\top}
\right]
\\
&=
\frac{1}{\gamma_0N}
\sum_i
\left[
\boldsymbol{w}_i(t)
\boldsymbol{a}_i^{\top}
\boldsymbol{R}(t)^{\top}
\boldsymbol{\Lambda}
+
\boldsymbol{\Lambda}
\boldsymbol{R}(t)
\boldsymbol{a}_i
\boldsymbol{w}_i(t)^{\top}
\right]
\\
&=
\frac{1}{\sigma_a^2}
\left[
\boldsymbol{G}^{wa}(t)
\dot{\boldsymbol{G}}^{wa}(t)^{\top}
+
\dot{\boldsymbol{G}}^{wa}(t)
\boldsymbol{G}^{wa}(t)^{\top}
\right],
\\
\therefore\qquad
\boldsymbol{G}^{w}(t)
-
\boldsymbol{G}^{w}(0)
&=
\frac{1}{\sigma_a^2}
\left[
\boldsymbol{G}^{wa}(t)
\boldsymbol{G}^{wa}(t)^{\top}
-
\boldsymbol{G}^{wa}(0)
\boldsymbol{G}^{wa}(0)^{\top}
\right].
\end{align}
In the limit \(N\to\infty\),
\begin{align}
\boldsymbol{G}^{wa}(0)
&=
\frac{1}{N}
\sum_i
\boldsymbol{w}_i(0)\boldsymbol{a}_i^{\top}
=
\boldsymbol{0},
\\
\boldsymbol{G}^{w}(0)
&=
\sigma_w^2\boldsymbol{I}.
\end{align}
Combining these results, we obtain
\begin{equation}
G^{w}_{kl}(t)
=
\sigma_w^2\delta_{kl}
+
\frac{1}{c}
\sum_p
\Theta_{kp}\Theta_{lp}
\left[
1-\exp(-c\lambda_kt)
\right]
\left[
1-\exp(-c\lambda_lt)
\right],
\end{equation}
where
\begin{equation}
c
=
\frac{\sigma_a^2}{\gamma_0^2}.
\end{equation}
Since
\begin{equation}
\sum_p\Theta_{kp}\Theta_{lp}
=
\mu_k\delta_{kl}
\end{equation}
by assumption, it follows that
\begin{equation}
G^{w}_{kl}(t)
=
\left\{
\sigma_w^2
+
\frac{\mu_k}{c}
\left[
1-\exp(-c\lambda_kt)
\right]^2
\right\}
\delta_{kl}.
\end{equation}
Therefore,
\begin{equation}
M_{kl}(t)
=
\left\{
\lambda_k\sigma_w^2
+
\frac{\lambda_k\mu_k}{c}
\left[
1-\exp(-c\lambda_kt)
\right]^2
\right\}
\delta_{kl},
\end{equation}
and the eigenvalues \(\rho_k(t)\) of \(\boldsymbol{M}(t)\) are
\begin{equation}
\rho_k(t)
=
\lambda_k\sigma_w^2
+
\frac{\lambda_k\mu_k}{c}
\left[
1-\exp(-c\lambda_kt)
\right]^2.
\end{equation}
The dynamics of this eigenvalue spectrum are illustrated in Fig.~\ref{fig:frozen-spectrum}.

\section{Warm-up: Derivation of the power-law exponent boundaries when the second layer is fixed}
\label{app:one-layer-exponent}
To characterize the power-law exponent around mode \(k\) at time \(t\), we introduce
\begin{equation}
\alpha_{\mathrm{eff}}(k,t)
:=
-
\frac{\partial\log\rho_k(t)}{\partial\log k}.
\end{equation}
We derive this quantity from the analytical expression for \(\rho_k(t)\). Substituting
\begin{equation}
\lambda_k=\lambda_0k^{-\alpha},
\qquad
\mu_k=\mu_0k^{-\beta},
\end{equation}
we obtain
\begin{align}
\rho_k(t)
&=
\lambda_0\sigma_w^2k^{-\alpha}
+
\frac{\lambda_0\mu_0k^{-(\alpha+\beta)}}{c}
\left[
1-\exp(-c\lambda_0tk^{-\alpha})
\right]^2
\\
&=
\lambda_0\sigma_w^2k^{-\alpha}
\left[
1
+
\frac{\mu_0k^{-\beta}}{c\sigma_w^2}
\left[
1-\exp(-c\lambda_0tk^{-\alpha})
\right]^2
\right]
\\
&=
\lambda_0\sigma_w^2k^{-\alpha}
\left[
1+Q(k,t)
\right],
\end{align}
where
\begin{align}
Q(k,t)
&=
\frac{\mu_0k^{-\beta}}{c\sigma_w^2}
\left[
1-\exp(-z_k(t))
\right]^2,
\\
z_k(t)
&=
c\lambda_0tk^{-\alpha}.
\end{align}
Taking logarithms gives
\begin{align}
\log\rho_k(t)
&=
\log(\lambda_0\sigma_w^2)
-
\alpha\log k
+
\log\!\left[1+Q(k,t)\right],
\\
\therefore\qquad
\alpha_{\mathrm{eff}}(k,t)
&=
\alpha
-
\frac{k}{1+Q(k,t)}
\frac{\partial Q(k,t)}{\partial k}.
\end{align}
The derivative of \(Q(k,t)\) is
\begin{align}
\frac{\partial Q(k,t)}{\partial k}
&=
-
\frac{\beta\mu_0k^{-\beta-1}}{c\sigma_w^2}
\left[
1-\exp(-z_k(t))
\right]^2
\\
&\quad
+
\frac{2\mu_0k^{-\beta}}{c\sigma_w^2}
\left[
1-\exp(-z_k(t))
\right]
\exp(-z_k(t))
\frac{\partial z_k(t)}{\partial k}
\\
&=
-
\frac{\mu_0k^{-\beta}}{c\sigma_w^2}
\left[
1-\exp(-z_k(t))
\right]^2
\left[
\frac{\beta}{k}
+
\frac{2}{\exp(z_k(t))-1}
\frac{\alpha z_k(t)}{k}
\right]
\\
&=
-
\frac{Q(k,t)}{k}
\left[
\beta
+
\frac{2\alpha z_k(t)}{\exp(z_k(t))-1}
\right].
\end{align}
Thus,
\begin{equation}
\alpha_{\mathrm{eff}}(k,t)
=
\alpha
+
\frac{Q(k,t)}{1+Q(k,t)}
\left[
\beta
+
\frac{2\alpha z_k(t)}{\exp(z_k(t))-1}
\right].
\end{equation}
This expression gives the analytical time evolution of the local power-law exponent. We next use it to derive approximate boundaries that partition the \((t,k)\) plane.

The time dependence of \(\rho_k(t)\) enters only through
\begin{equation}
z_k(t)
=
c\lambda_0tk^{-\alpha}.
\end{equation}
Specifically, since \(Q(k,t)\) is a sigmoidal function with respect to \(t\), a mode \(k\) is unlearned at time \(t\) when \(z_k(t)\ll1\), and learned when \(z_k(t)\gg1\). Defining
\begin{equation}
k_{\mathrm{learn}}(t)
=
(c\lambda_0t)^{1/\alpha},
\end{equation}
the mode is unlearned for \(k\gg k_{\mathrm{learn}}(t)\) and learned for \(k\ll k_{\mathrm{learn}}(t)\). On a log--log plot, the boundary between the learned and unlearned regions is the line
\begin{equation}
\log k_{\mathrm{learn}}(t)
=
\frac{1}{\alpha}
\left[
\log t+\log(c\lambda_0)
\right].
\end{equation}
We now obtain the asymptotic power-law exponents in each region .

\subsection{Learned modes}
\label{app:fixed-second-layer-learned}

For learned modes satisfying \(k\ll k_{\mathrm{learn}}(t)\), we have \(z_k(t)\gg1\), and hence
\begin{equation}
Q(k,t)
\simeq
\frac{\mu_0k^{-\beta}}{c\sigma_w^2}.
\end{equation}
Therefore,
\begin{equation}
\rho_k(t)
\simeq
\lambda_0\sigma_w^2k^{-\alpha}
+
\frac{\lambda_0\mu_0}{c}
k^{-(\alpha+\beta)}.
\end{equation}
The first term is the input-induced anisotropy present at initialization, whereas the second is the anisotropy induced by learning. Because the second term decays more rapidly with \(k\), the spectrum must ultimately decay with exponent \(\alpha\) at sufficiently large \(k\). This need not hold, however, at finite \(k\). In particular, for
\begin{equation}
k
\ll
k_1
:=
\left(
\frac{\mu_0}{c\sigma_w^2}
\right)^{1/\beta},
\end{equation}
the spectrum decays with exponent \(\alpha+\beta\). On a log--log plot, this boundary is the constant line
\begin{equation}
\log k_1
=
\frac{1}{\beta}
\log
\left(
\frac{\mu_0}{c\sigma_w^2}
\right).
\end{equation}

\subsection{Unlearned modes}
\label{app:fixed-second-layer-unlearned}

For \(k\gg k_{\mathrm{learn}}(t)\), we have
\begin{equation}
Q(k,t)
\simeq
\frac{\mu_0k^{-\beta}z_k(t)^2}{c\sigma_w^2}.
\end{equation}
It follows that
\begin{align}
\rho_k(t)
&\simeq
\lambda_0\sigma_w^2k^{-\alpha}
+
\frac{\lambda_0\mu_0k^{-(\alpha+\beta)}}{c}
z_k(t)^2
\\
&=
\lambda_0\sigma_w^2k^{-\alpha}
+
c\lambda_0^3\mu_0t^2k^{-(3\alpha+\beta)}.
\end{align}
As in the learned case, the second term decays more rapidly with \(k\), so the asymptotic exponent at sufficiently large \(k\) is necessarily \(\alpha\). At finite \(k\), however, modes satisfying
\begin{equation}
k
\ll
k_2(t)
:=
\left(
\frac{c\lambda_0^2\mu_0}{\sigma_w^2}
t^2
\right)^{1/(2\alpha+\beta)}
\end{equation}
decay with exponent \(3\alpha+\beta\). On a log--log plot, this boundary is the line
\begin{equation}
\log k_2(t)
=
\frac{1}{2\alpha+\beta}
\left[
2\log t
+
\log
\left(
\frac{c\lambda_0^2\mu_0}{\sigma_w^2}
\right)
\right].
\end{equation}

\begin{figure}
  \centering
  \includegraphics[width=0.7\linewidth]{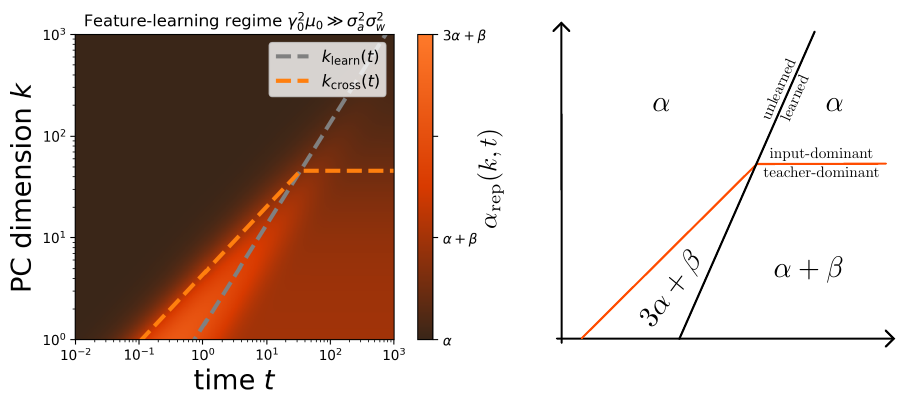}
  \caption{Asymptotic exponent regions for the model with a fixed second layer.}
  \label{fig:frozen-regions}
\end{figure}

From the two cases above, the intersection of \(k_1\) and \(k_2(t)\) is
\begin{equation}
(\log t^*,\log k^*)
=
\left(
\frac{\alpha}{\beta}
\log
\frac{\mu_0}{c\sigma_w^2}
-
\log(c\lambda_0),
\;
\frac{1}{\beta}
\log
\frac{\mu_0}{c\sigma_w^2}
\right),
\end{equation}
which lies on \(k_{\mathrm{learn}}(t)\). Defining
\begin{equation}
k_{\mathrm{cross}}(t)
=
\begin{cases}
k_1,
& t>t^*,
\\
k_2(t),
& t\leq t^*,
\end{cases}
\end{equation}
we can summarize the asymptotic behavior as 
\begin{equation}
\alpha_{\mathrm{eff}}(k,t)
\simeq
\begin{cases}
\alpha,
& k>k_{\mathrm{cross}}(t),
\\
3\alpha+\beta,
& k_{\mathrm{learn}}(t)<k<k_{\mathrm{cross}}(t),
\\
\alpha+\beta,
& \text{otherwise}.
\end{cases}
\end{equation}
These three regions and their asymptotic exponents are illustrated in Fig.~\ref{fig:frozen-regions}.

\section{Analytical eigenvalue solution when both layers are trained}
\label{app:both-layers-solution}

The network output is 
\begin{align}
\boldsymbol{f}(\boldsymbol{x},t)
&=
\frac{1}{\gamma_0N}
\sum_{i=1}^{N}
\boldsymbol{a}_i(t)
\boldsymbol{w}_i(t)^{\top}
\boldsymbol{x}
\\
&=
\frac{1}{\gamma_0}
\boldsymbol{G}^{wa}(t)^{\top}
\boldsymbol{x}.
\end{align}
Thus, in the population limit, using
\begin{equation}
\boldsymbol{R}(t)
=
\boldsymbol{\Theta}
-
\frac{1}{\gamma_0}
\boldsymbol{G}^{wa}(t),
\end{equation}
the loss can be written as
\begin{align}
L(t)
&=
\frac{1}{2}
\mathbb{E}_{\boldsymbol{x}}
\left[
\left\lVert
\boldsymbol{f}(\boldsymbol{x},t)
-
\boldsymbol{y}(\boldsymbol{x})
\right\rVert^2
\right]
\\
&=
\frac{1}{2}
\operatorname{Tr}
\left[
\boldsymbol{R}(t)^{\top}
\boldsymbol{\Lambda}
\boldsymbol{R}(t)
\right].
\end{align}
The gradient-flow dynamics are therefore
\begin{align}
\dot{w}_{ik}(t)
&=
-\nu
\frac{\partial L}{\partial w_{ik}}
\\
&=
-\nu
\sum_{p=1}^{P}
\frac{\partial L}{\partial G^{wa}_{kp}}
\frac{\partial G^{wa}_{kp}}{\partial w_{ik}}
\\
&=
\frac{\nu}{\gamma_0N}
\sum_{p=1}^{P}
\lambda_kR_{kp}(t)a_{ip}(t)
\\
&=
\frac{1}{\gamma_0}
\sum_{p=1}^{P}
\lambda_kR_{kp}(t)a_{ip}(t),
\\
\therefore\qquad
\dot{\boldsymbol{w}}_i(t)
&=
\frac{1}{\gamma_0}
\boldsymbol{\Lambda}
\boldsymbol{R}(t)
\boldsymbol{a}_i(t),
\end{align}
and
\begin{align}
\dot{a}_{ip}(t)
&=
-\nu
\frac{\partial L}{\partial a_{ip}}
\\
&=
-\nu
\sum_{k=1}^{D}
\frac{\partial L}{\partial G^{wa}_{kp}}
\frac{\partial G^{wa}_{kp}}{\partial a_{ip}}
\\
&=
\frac{\nu}{\gamma_0N}
\sum_{k=1}^{D}
\lambda_kR_{kp}(t)w_{ik}(t)
\\
&=
\frac{1}{\gamma_0}
\sum_{k=1}^{D}
\lambda_kR_{kp}(t)w_{ik}(t),
\\
\therefore\qquad
\dot{\boldsymbol{a}}_i(t)
&=
\frac{1}{\gamma_0}
\boldsymbol{R}(t)^{\top}
\boldsymbol{\Lambda}
\boldsymbol{w}_i(t).
\end{align}
Here again, we set \(\nu=N\).

These weight dynamics yield the following Gram-matrix dynamics:
\begin{align}
\dot{\boldsymbol{G}}^{wa}(t)
&=
\frac{1}{N}
\sum_{i=1}^{N}
\left[
\dot{\boldsymbol{w}}_i(t)
\boldsymbol{a}_i(t)^{\top}
+
\boldsymbol{w}_i(t)
\dot{\boldsymbol{a}}_i(t)^{\top}
\right]
\\
&=
\frac{1}{\gamma_0N}
\sum_{i=1}^{N}
\left[
\boldsymbol{\Lambda}
\boldsymbol{R}(t)
\boldsymbol{a}_i(t)
\boldsymbol{a}_i(t)^{\top}
+
\boldsymbol{w}_i(t)
\boldsymbol{w}_i(t)^{\top}
\boldsymbol{\Lambda}
\boldsymbol{R}(t)
\right]
\\
&=
\frac{1}{\gamma_0}
\left[
\boldsymbol{\Lambda}
\boldsymbol{R}(t)
\boldsymbol{G}^{a}(t)
+
\boldsymbol{G}^{w}(t)
\boldsymbol{\Lambda}
\boldsymbol{R}(t)
\right],
\end{align}
\begin{align}
\dot{\boldsymbol{G}}^{w}(t)
&=
\frac{1}{N}
\sum_{i=1}^{N}
\left[
\dot{\boldsymbol{w}}_i(t)
\boldsymbol{w}_i(t)^{\top}
+
\boldsymbol{w}_i(t)
\dot{\boldsymbol{w}}_i(t)^{\top}
\right]
\\
&=
\frac{1}{\gamma_0N}
\sum_{i=1}^{N}
\left[
\boldsymbol{\Lambda}
\boldsymbol{R}(t)
\boldsymbol{a}_i(t)
\boldsymbol{w}_i(t)^{\top}
+
\boldsymbol{w}_i(t)
\boldsymbol{a}_i(t)^{\top}
\boldsymbol{R}(t)^{\top}
\boldsymbol{\Lambda}
\right]
\\
&=
\frac{1}{\gamma_0}
\left[
\boldsymbol{\Lambda}
\boldsymbol{R}(t)
\boldsymbol{G}^{wa}(t)^{\top}
+
\boldsymbol{G}^{wa}(t)
\boldsymbol{R}(t)^{\top}
\boldsymbol{\Lambda}
\right],
\end{align}
and
\begin{align}
\dot{\boldsymbol{G}}^{a}(t)
&=
\frac{1}{N}
\sum_{i=1}^{N}
\left[
\dot{\boldsymbol{a}}_i(t)
\boldsymbol{a}_i(t)^{\top}
+
\boldsymbol{a}_i(t)
\dot{\boldsymbol{a}}_i(t)^{\top}
\right]
\\
&=
\frac{1}{\gamma_0N}
\sum_{i=1}^{N}
\left[
\boldsymbol{R}(t)^{\top}
\boldsymbol{\Lambda}
\boldsymbol{w}_i(t)
\boldsymbol{a}_i(t)^{\top}
+
\boldsymbol{a}_i(t)
\boldsymbol{w}_i(t)^{\top}
\boldsymbol{\Lambda}
\boldsymbol{R}(t)
\right]
\\
&=
\frac{1}{\gamma_0}
\left[
\boldsymbol{R}(t)^{\top}
\boldsymbol{\Lambda}
\boldsymbol{G}^{wa}(t)
+
\boldsymbol{G}^{wa}(t)^{\top}
\boldsymbol{\Lambda}
\boldsymbol{R}(t)
\right].
\end{align}

Because the weights are initialized independently and \(N\to\infty\),
\begin{equation}
\boldsymbol{G}^{wa}(0)=\boldsymbol{0},
\qquad
\boldsymbol{G}^{w}(0)=\sigma_w^2\boldsymbol{I},
\qquad
\boldsymbol{G}^{a}(0)=\sigma_a^2\boldsymbol{I}.
\end{equation}
Moreover, by assumption, \(\boldsymbol{\Lambda}\) and \(\boldsymbol{\Theta}\boldsymbol{\Theta}^{\top}\) are diagonal. Since an orthogonal rotation in the output space leaves the loss invariant, we may set, without loss of generality,
\begin{equation}
\boldsymbol{\Theta}
=
\operatorname{diag}(\theta_0,\theta_1,\ldots)
=
\operatorname{diag}(\sqrt{\mu_0},\sqrt{\mu_1},\ldots).
\end{equation}
The derived Gram-matrix dynamics shows that if
\(\boldsymbol{\Lambda}\), \(\boldsymbol{\Theta}\),
\(\boldsymbol{G}^{wa}\), \(\boldsymbol{G}^{w}\), and
\(\boldsymbol{G}^{a}\) are all diagonal, then their time derivatives are also diagonal. Because they are diagonal at \(t=0\), it follows that
\(\boldsymbol{G}^{wa}(t)\), \(\boldsymbol{G}^{w}(t)\), and
\(\boldsymbol{G}^{a}(t)\) remain diagonal for all \(t\).

It therefore suffices to consider the independent dynamics of the diagonal entries. Define
\begin{align}
\xi_k(t)
&=
G^{wa}_{kk}(t),
\\
\eta_k(t)
&=
G^{w}_{kk}(t),
\\
\zeta_k(t)
&=
G^{a}_{kk}(t).
\end{align}
The initial conditions are
\begin{equation}
\xi_k(0)=0,
\qquad
\eta_k(0)=\sigma_w^2,
\qquad
\zeta_k(0)=\sigma_a^2,
\end{equation}
and the equations of motion are
\begin{align}
\dot{\xi}_k(t)
&=
\frac{\lambda_k}{\gamma_0}
\left(
\theta_k-\frac{\xi_k(t)}{\gamma_0}
\right)
\left[
\eta_k(t)+\zeta_k(t)
\right],
\\
\dot{\eta}_k(t)
=
\dot{\zeta}_k(t)
&=
\frac{2\lambda_k\xi_k(t)}{\gamma_0}
\left(
\theta_k-\frac{\xi_k(t)}{\gamma_0}
\right).
\end{align}
We seek the eigenvalues of
\begin{equation}
\boldsymbol{M}(t)
=
\boldsymbol{\Lambda}^{1/2}
\boldsymbol{G}^{w}(t)
\boldsymbol{\Lambda}^{1/2}.
\end{equation}
In the present setting, these are
\begin{equation}
\rho_k(t)
=
\lambda_k\eta_k(t).
\end{equation}

The system has two conserved quantities, allowing us to eliminate two variables. First, because
\(\dot{\eta}_k-\dot{\zeta}_k=0\),
\begin{equation}
\eta_k(t)-\zeta_k(t)
=
\sigma_w^2-\sigma_a^2
\qquad
\text{(constant)}.
\end{equation}
Next, define
\begin{equation}
s_k(t)
:=
\eta_k(t)+\zeta_k(t).
\end{equation}
Then
\begin{align}
\dot{s}_k
&=
\frac{4\lambda_k\xi_k}{\gamma_0}
\left(
\theta_k-\frac{\xi_k}{\gamma_0}
\right),
\\
\frac{ds_k}{d\xi_k}
&=
\frac{\dot{s}_k}{\dot{\xi}_k}
=
\frac{4\xi_k}{s_k},
\\
s_k\,ds_k
&=
4\xi_k\,d\xi_k,
\\
s_k(t)^2-s_k(0)^2
&=
4\xi_k(t)^2-4\xi_k(0)^2,
\\
s_k(t)^2-4\xi_k(t)^2
&=
(\sigma_w^2+\sigma_a^2)^2
\qquad
\text{(constant)}.
\end{align}
Using these two conserved quantities, \(\eta_k(t)\) can be expressed as a function of \(\xi_k(t)\):
\begin{equation}
\left\{
\begin{aligned}
\eta_k(t)+\zeta_k(t)
&=
\sqrt{
4\xi_k(t)^2
+
(\sigma_w^2+\sigma_a^2)^2
},
\\
\eta_k(t)-\zeta_k(t)
&=
\sigma_w^2-\sigma_a^2,
\end{aligned}
\right.
\end{equation}
and hence
\begin{equation}
\eta_k(t)
=
\sigma_w^2
+
\frac{1}{2}
\left[
-(\sigma_w^2+\sigma_a^2)
+
\sqrt{
4\xi_k(t)^2
+
(\sigma_w^2+\sigma_a^2)^2
}
\right].
\end{equation}
Thus, solving
\begin{equation}
\dot{\xi}_k
=
\frac{\lambda_k}{\gamma_0}
\left(
\theta_k-\frac{\xi_k}{\gamma_0}
\right)
\sqrt{
4\xi_k(t)^2
+
(\sigma_w^2+\sigma_a^2)^2
}
\label{eq:xi-difeq}
\end{equation}
allows us to obtain the analytical eigenvalue solution through the sequence
\(\xi_k(t)\to \eta_k(t)\to\rho_k(t)\).

We solve for \(\xi_k(t)\) by a change of variables and separation of variables. Let
\begin{equation}
S
:=
\sigma_w^2+\sigma_a^2,
\quad
m_k
:=
\gamma_0\theta_k,
\quad
b_k
:=
\frac{\lambda_k}{\gamma_0^2},
\quad
u_k := \frac{\xi}{S},
\quad
\ell_k
:=
\operatorname{arsinh}
\left(2u_k
\right).
\end{equation}
Then
\begin{equation}
\left\{
\begin{aligned}
\text{left-hand side}
&=
\dot{\xi}_k
=
\frac{S}{2}
\cosh(\ell_k)\dot{\ell}_k,
\\
\text{right-hand side}
&=
b_k
\left(
m_k-\frac{S}{2}\sinh(\ell_k)
\right)
S\cosh(\ell_k).
\end{aligned}
\right.
\end{equation}
The initial-value problem is therefore
\begin{equation}
\dot{\ell}_k
=
b_k
\left(
2m_k-S\sinh(\ell_k)
\right),
\qquad
\ell_k(0)=0.
\end{equation}
Next, let
\begin{equation}
v_k
=
\tanh
\left(
\frac{\ell_k}{2}
\right).
\end{equation}
Then
\begin{equation}
\left\{
\begin{aligned}
\dot{v}_k
&=
\frac{\dot{\ell}_k}
{2\cosh^2(\ell_k/2)}
=
\frac{1-v_k^2}{2}
\dot{\ell}_k,
\\
\sinh(\ell_k)
&=
\frac{2v_k}{1-v_k^2}.
\end{aligned}
\right.
\end{equation}
It follows that
\begin{equation}
\dot{v}_k
=
b_k
\left[
m_k(1-v_k^2)-Sv_k
\right],
\qquad
v_k(0)=0.
\end{equation}
This equation can be solved by partial fractions. Let
\begin{equation}
r_k^{\pm}
=
\frac{
-S\pm\sqrt{S^2+4m_k^2}
}
{2m_k}
\end{equation}
be the two roots of
\begin{equation}
m_kv_k^2+Sv_k-m_k=0.
\end{equation}
Then
\begin{align}
b_kt
&=
\frac{1}{m_k(r_k^+-r_k^-)}
\int_0^{v_k(t)}
\left(
\frac{1}{v-r_k^-}
-
\frac{1}{v-r_k^+}
\right)
\,dv
\\
&=
\frac{1}{m_k(r_k^+-r_k^-)}
\log
\left|
\frac{
r_k^+\left(v_k(t)-r_k^-\right)
}{
r_k^-\left(v_k(t)-r_k^+\right)
}
\right|
\\
&=
\frac{1}{\sqrt{S^2+4m_k^2}}
\log
\left|
\frac{
r_k^+\left(v_k(t)-r_k^-\right)
}{
r_k^-\left(v_k(t)-r_k^+\right)
}
\right|.
\end{align}
This equation can be inverted to obtain \(v_k(t)\). Define
\begin{equation}
E_k(t)
:=
\exp
\left[
b_k\sqrt{S^2+4m_k^2}\,t
\right].
\end{equation}
Then
\begin{align}
\frac{
r_k^+\left(v_k(t)-r_k^-\right)
}{
r_k^-\left(v_k(t)-r_k^+\right)
}
&=
E_k(t),
\\
\therefore\qquad
v_k(t)
&=
\frac{
r_k^+r_k^-\left[1-E_k(t)\right]
}{
r_k^+-r_k^-E_k(t)
}
\\
&=
\frac{
r_k^+\left[E_k(t)-1\right]
}{
(r_k^+)^2+E_k(t)
},
\end{align}
where we used \(r_k^+r_k^-=-1\) in the last line.

The analytical solution can therefore be expressed through the sequence
\(E_k(t)\to v_k(t)\to u_k(t)\to \eta_k(t)\):
\begin{align}
\eta_k(t)
&=
\sigma_w^2
+
\frac{S}{2}
\left[
-1
+
\sqrt{1+ 4u_k(t)^2}
\right],
\\
u_k(t)
&=
\frac{v_k(t)}{1-v_k(t)^2},
\\
v_k(t)
&=
\frac{
r_k^+\left[E_k(t)-1\right]
}{
(r_k^+)^2+E_k(t)
},
\\
E_k(t)
&=
\exp
\left[
\frac{
\lambda_k
\sqrt{S^2+4\gamma_0^2\theta_k^2}
}{
\gamma_0^2
}
t
\right],
\\
r_k^+
&=
\frac{
-S
+
\sqrt{S^2+4\gamma_0^2\theta_k^2}
}{
2\gamma_0\theta_k
},
\\
S
&=
\sigma_w^2+\sigma_a^2.
\end{align}
This provides the analytical solution for the eigenvalues \(\rho_k(t)=\lambda_k\eta_k(t)\). As explained in the next section, however, the full analytical solution is not required to evaluate the power-law exponents.

\section{Derivation of the power-law exponents when both layers are trained}
\label{app:boundaries}

\subsection{Exact local exponent}
We derive the analytical expression for the power-law exponent presented in Result 3.2. The derivation uses the following equations:

\[
\left\{
\begin{aligned}
\rho_k(t)
&=\lambda_k\sigma_w^2\left[1+R(u_k)\right],\\
R(u_k)
&=\frac{S}{2\sigma_w^2}\left[-1+g(u_k)\right],\\
\dot{u}_k
&=\frac{\lambda_k S}{\gamma_0^2}
(u_{k,\infty}-u_k)g(u_k),
\end{aligned}
\right.
\]

with the initial condition \(u_k(0)=0\), where

\[
u_{k,\infty}:=\frac{\gamma_0\theta_k}{S},
\qquad
S:=\sigma_w^2+\sigma_a^2,
\qquad
g(x):=\sqrt{1+4x^2}.
\]

We first define

\[
p(k,t):=-\frac{\partial\log u_k(t)}{\partial\log k}
\]

and express \(\alpha_{\mathrm{eff}}(k,t)\) in terms of \(p(k,t)\). From the analytical expression for the eigenspectrum, we obtain

\[
\begin{aligned}
\alpha_{\mathrm{eff}}(k,t)
&=-\frac{\partial\log\rho_k(t)}{\partial\log k}\\
&=\alpha+
\frac{R(u_k(t))}{1+R(u_k(t))}
\left(
-\frac{\partial\log R(u_k(t))}{\partial\log k}
\right)\\
&=\alpha+
\frac{R(u_k(t))}{1+R(u_k(t))}
\left(1+\frac{1}{g(u_k(t))}\right)
\left(
-\frac{\partial\log u_k(t)}{\partial\log k}
\right)\\
&=\alpha+
\frac{R(u_k(t))}{1+R(u_k(t))}
\left(1+\frac{1}{g(u_k(t))}\right)p(k,t).
\end{aligned}
\]

Next, we determine \(p(k,t)\) from the evolution equation for \(u_k(t)\). Separating variables and integrating yields

\[
\frac{\lambda_k St}{\gamma_0^2}
=F(u_k(t),u_{k,\infty})
:=\int_0^{u_k(t)}
\frac{dz}{(u_{k,\infty}-z)g(z)}.
\]

Differentiating both sides with respect to \(\log k\) at fixed \(t\), the left-hand side becomes

\[
\begin{aligned}
\frac{\partial}{\partial\log k}
\frac{\lambda_k St}{\gamma_0^2}
&=k\frac{\partial}{\partial k}
\frac{\lambda_k St}{\gamma_0^2}\\
&=-\alpha\frac{\lambda_k St}{\gamma_0^2}\\
&=-\alpha F(u_k(t),u_{k,\infty}),
\end{aligned}
\]

while the right-hand side becomes

\[
\frac{\partial F}{\partial\log k}
=
u_k(t)\frac{\partial F}{\partial u_k(t)}
\frac{\partial\log u_k(t)}{\partial\log k}
+
\frac{\partial F}{\partial u_{k,\infty}}
\frac{\partial u_{k,\infty}}{\partial\log k}.
\]

Evaluating the derivatives and rearranging, we obtain

\[
p(k,t)
=
\frac{(u_{k,\infty}-u_k(t))g(u_k(t))}{u_k(t)}
\int_0^{u_k(t)}
\left[
\frac{\alpha}{u_{k,\infty}-z}
+\frac{u_{k,\infty}\beta/2}{(u_{k,\infty}-z)^2}
\right]\frac{dz}{g(z)}.
\]

\subsection{Boundaries}
Since \(u_k(t)\) increases monotonically from zero toward \(u_{k,\infty}=\gamma_0\theta_k/S\), we can define the hitting time \(t_k^*(u)\) for each threshold \(0<u<u_{k,\infty}\):
\begin{equation}
t_k^*(u)
=
\frac{\gamma_0^2}{\lambda_kS}
\int_0^u
\frac{
dz
}{
\left(
u_{k,\infty}-z
\right)
\sqrt{
1+4z^2
}
}.
\end{equation}
The boundaries \(k_{\mathrm{cross}}\), \(k_{\mathrm{learn}}\), and \(k_{\mathrm{layer}}\) are then defined implicitly by
\[
t=t_{k_{\mathrm{cross}}(t)}^*(u_{\mathrm{cross}}),
\qquad
t=t_{k_{\mathrm{learn}}(t)}^*(0.9u_{k,\infty}),
\qquad
t=t_{k_{\mathrm{layer}}(t)}^*(1),
\]whenever the corresponding thresholds are reachable. The uniqueness of these boundaries follows from the preservation of the ordering of \(u_k(t)\) across modes throughout the dynamics, as shown in Appendix \ref{app:ordering}.

\subsection{Exponents within each domain}
\label{app:derivation-exponents}
We first show that
\begin{equation}
0 \le
p(k,t) -
\left[
\alpha h\left(\frac{u_k(t)}{u_{k,\infty}}\right)
+\frac{\beta}{2}
\right]
\le
2\left(\alpha+\frac{\beta}{2}\right)u_k(t)^2,
\tag{*}\label{eq:p-bound}
\end{equation}
holds for $t\ge0$, where
\[
h(x):=\frac{1-x}{x}\log\frac{1}{1-x}.
\]

Using the fact that
\[
1 \le \frac{g(u_k(t))}{g(z)} \le g(u_k(t))
\]
throughout the integration interval in the exact expression,
we obtain
\begin{align*}
p(k,t)
&\ge
\frac{u_{k,\infty}-u_k(t)}{u_k(t)}
\int_0^{u_k(t)}
\left[
\frac{\alpha}{u_{k,\infty}-z}
+\frac{u_{k,\infty}\beta/2}{(u_{k,\infty}-z)^2}
\right]\,dz \\
&=
\alpha h\left(\frac{u_k(t)}{u_{k,\infty}}\right)
+\frac{\beta}{2}, \\
p(k,t)
&\le
g(u_k(t))
\left[
\alpha h\left(\frac{u_k(t)}{u_{k,\infty}}\right)
+\frac{\beta}{2}
\right].
\end{align*}

Furthermore,
\begin{align*}
&p(k,t) -
\left[
\alpha h\left(\frac{u_k(t)}{u_{k,\infty}}\right)
+\frac{\beta}{2}
\right] \\
&\quad \le
\bigl(g(u_k(t))-1\bigr)
\left[
\alpha h\left(\frac{u_k(t)}{u_{k,\infty}}\right)
+\frac{\beta}{2}
\right] \\
&\quad \le
\bigl(g(u_k(t))-1\bigr)
\left(\alpha+\frac{\beta}{2}\right)
\qquad \bigl(h(x)\le 1\bigr) \\
&\quad \le
2u_k(t)^2\left(\alpha+\frac{\beta}{2}\right)
\qquad \bigl(g(x)\le 1+2x^2\bigr).
\end{align*}
This establishes inequality~(\ref{eq:p-bound}).
We now evaluate the power-law exponent in each regime.

\paragraph{Region I.}
We bound each factor in the exact expression
\[
\alpha_{\mathrm{eff}}(k,t)
=
\alpha+
\frac{R(u_k(t))}{1+R(u_k(t))}
\left(1+\frac{1}{g(u_k(t))}\right)p(k,t).
\]
First,
\begin{align*}
\frac{R(u_k(t))}{1+R(u_k(t))}
\left(1+\frac{1}{g(u_k(t))}\right)
&\le 2R(u_k(t))
&& \bigl(g(u)\ge1,\ R(u)\ge0\bigr) \\
&\le \frac{2S}{\sigma_w^2}u_k(t)^2
&& \bigl(g(x)\le1+2x^2\bigr) \\
&\le 4\left(\frac{u_k(t)}{u_{\mathrm{cross}}}\right)^2,
\end{align*}
where the last inequality follows from
\[
u_{\mathrm{cross}}^2
=\frac{\sigma_w^2(S+\sigma_w^2)}{S^2}
\qquad\text{and}\qquad
S\ge\sigma_w^2.
\]

Moreover, inequality~(\ref{eq:p-bound}), together with
$h(x)\le1$, gives
\[
0\le p(k,t)
\le
\left(\alpha+\frac{\beta}{2}\right)
\left(1+2u_k(t)^2\right)
=O(1)
\]
in Region I. Consequently,
\[
\alpha_{\mathrm{eff}}(k,t)
=
\alpha+
O\left(
\left(\frac{u_k(t)}{u_{\mathrm{cross}}}\right)^2
\right).
\]

\paragraph{Region II.}
As above, we estimate each factor separately. In the regime
$u_{\mathrm{cross}}\ll u_k(t)\ll\min(1,u_{k,\infty})$,
we have
\begin{align*}
&\frac{R(u_k(t))}{1+R(u_k(t))}
\left(1+\frac{1}{g(u_k(t))}\right) \\
&\quad =
\left(1+O\left(R(u_k(t))^{-1}\right)\right)
\left(2+O\left(u_k(t)^2\right)\right) \\
&\quad =
2+O\left(
\left(\frac{u_{\mathrm{cross}}}{u_k(t)}\right)^2
+u_k(t)^2
\right).
\end{align*}

By inequality~(\ref{eq:p-bound}),
\begin{align*}
p(k,t)
&=
\left[
\alpha h\left(\frac{u_k(t)}{u_{k,\infty}}\right)
+\frac{\beta}{2}
\right]
+O\left(u_k(t)^2\right) \\
&=
\alpha+\frac{\beta}{2}
+O\left(
\frac{u_k(t)}{u_{k,\infty}}+u_k(t)^2
\right),
\end{align*}
where we used $h(x)=1+O(x)$ as $x\to0$.

Consequently,
\[
\alpha_{\mathrm{eff}}(k,t)
=
3\alpha+\beta
+O\left(
\left(\frac{u_{\mathrm{cross}}}{u_k(t)}\right)^2
+\frac{u_k(t)}{u_{k,\infty}}
+u_k(t)^2
\right).
\]
By contrast, in a domain where \(1 \ll u_k(t)\ll u_{k,\infty}\), \(p(k,t)\) remains mode-dependent. The eigenvalue
spectrum \(\rho_k(t)\) then has no single power-law exponent.
\paragraph{Region IIIa.}
As in Region II, the first factor satisfies
\[
\frac{R(u_k(t))}{1+R(u_k(t))}
\left(1+\frac{1}{g(u_k(t))}\right)
=
2+O\left(
\left(\frac{u_{\mathrm{cross}}}{u_k(t)}\right)^2
+u_k(t)^2
\right).
\]

For the second factor, inequality~(\ref{eq:p-bound}) gives
\begin{align*}
p(k,t)
&=
\alpha h\left(\frac{u_k(t)}{u_{k,\infty}}\right)
+\frac{\beta}{2}
+O\left(u_k(t)^2\right) \\
&\to
\frac{\beta}{2}+O\left(u_k(t)^2\right)
\qquad \bigl(u_k(t)\to u_{k,\infty}\bigr),
\end{align*}
since $h(x)\to0$ as $x\to1^{-}$.

Consequently,
\[
\alpha_{\mathrm{eff}}(k,t)
\to
\alpha+\beta
+O\left(
\left(\frac{u_{\mathrm{cross}}}{u_k(t)}\right)^2
+u_k(t)^2
\right).
\]

\paragraph{Region IIIb.}
For the first factor, we have
\begin{align*}
&\frac{R(u_k(t))}{1+R(u_k(t))}
\left(1+\frac{1}{g(u_k(t))}\right) \\
&\quad =
\left(1+O\left(R(u_k(t))^{-1}\right)\right)
\left(1+O\left(u_k(t)^{-1}\right)\right) \\
&\quad =
1+O\left(
\left(\frac{u_{\mathrm{cross}}}{u_k(t)}\right)^2
+u_k(t)^{-1}
\right).
\end{align*}

For the second factor, the bound in
inequality~(\ref{eq:p-bound}) is not sufficiently sharp,
since $u_k(t)\gg1$ in this regime.
We therefore use integration by parts to obtain a sharper bound.

Integrating both sides of the identity
\[
\frac{d}{dz}
\frac{z}{(u_{k,\infty}-z)g(z)}
=
\frac{u_{k,\infty}}{(u_{k,\infty}-z)^2g(z)}
-
\frac{4z^2}{(u_{k,\infty}-z)g(z)^3}
\]
gives
\begin{align*}
&\frac{u_k(t)}
{(u_{k,\infty}-u_k(t))g(u_k(t))} \\
&\quad =
\int_0^{u_k(t)}
\frac{u_{k,\infty}\,dz}
{(u_{k,\infty}-z)^2g(z)}
-
\int_0^{u_k(t)}
\frac{4z^2}{1+4z^2}
\frac{dz}{(u_{k,\infty}-z)g(z)}.
\end{align*}
Substituting this identity into the exact expression
for $p(k,t)$ yields
\begin{align*}
p(k,t)
&=
\frac{\beta}{2}
+
\frac{(u_{k,\infty}-u_k(t))g(u_k(t))}{u_k(t)}
\\
&\qquad \times
\int_0^{u_k(t)}
\frac{\alpha+\frac{\beta}{2}\frac{4z^2}{1+4z^2}}
{(u_{k,\infty}-z)g(z)}\,dz.
\end{align*}
Using
\[
0\le\frac{4z^2}{1+4z^2}\le1,
\]
we obtain
\begin{align*}
0\le p(k,t)-\frac{\beta}{2}
&\le
\left(\alpha+\frac{\beta}{2}\right)
\frac{(u_{k,\infty}-u_k(t))g(u_k(t))}{u_k(t)}
\\
&\qquad \times
\int_0^{u_k(t)}
\frac{dz}{(u_{k,\infty}-z)g(z)}.
\end{align*}

To estimate the remaining integral, we use the decomposition
\begin{align*}
&\frac{u_{k,\infty}g(u_k(t))}{u_k(t)}
\int_0^{u_k(t)}
\frac{dz}{(u_{k,\infty}-z)g(z)} \\
&\quad =
\frac{g(u_k(t))}{u_k(t)}
\int_0^{u_k(t)}\frac{dz}{g(z)}
+
\int_0^{u_k(t)}
\frac{g(u_k(t))}{u_k(t)}
\frac{z}{g(z)}
\frac{dz}{u_{k,\infty}-z}.
\end{align*}
The first integral satisfies
\[
\int_0^{u_k(t)}\frac{dz}{g(z)}
=
\frac{\operatorname{arsinh}(2u_k(t))}{2}
=
O\left(\log(4u_k(t))\right).
\]
Moreover, since $z/g(z)$ is increasing for $z\ge0$,
\[
\frac{g(u_k(t))}{u_k(t)}\frac{z}{g(z)}
\le1
\qquad (0\le z\le u_k(t)).
\]
Thus,
\begin{align*}
\int_0^{u_k(t)}
\frac{g(u_k(t))}{u_k(t)}
\frac{z}{g(z)}
\frac{dz}{u_{k,\infty}-z}
&\le
\int_0^{u_k(t)}
\frac{dz}{u_{k,\infty}-z} \\
&=
\log\frac{u_{k,\infty}}{u_{k,\infty}-u_k(t)}.
\end{align*}
Combining these estimates with
$g(u_k(t))/u_k(t)=O(1)$ for $u_k(t)\gg1$, we obtain
\begin{align*}
&\frac{(u_{k,\infty}-u_k(t))g(u_k(t))}{u_k(t)}
\int_0^{u_k(t)}
\frac{dz}{(u_{k,\infty}-z)g(z)} \\
&\quad =
O\left(
\frac{u_{k,\infty}-u_k(t)}{u_{k,\infty}}
\log\frac{4u_k(t)u_{k,\infty}}
{u_{k,\infty}-u_k(t)}
\right).
\end{align*}
Consequently,
\[
p(k,t)
=
\frac{\beta}{2}
+
O\left(
\frac{u_{k,\infty}-u_k(t)}{u_{k,\infty}}
\log\frac{4u_k(t)u_{k,\infty}}
{u_{k,\infty}-u_k(t)}
\right),
\]
with $p(k,t)\ge\beta/2$.

For any fixed \(k\), the error term vanishes as
$u_k(t)\to u_{k,\infty}^{-}$. at sufficiently large \(t\). Consequently, sufficiently far to the right within Region IIIb,
\[
\alpha_{\mathrm{eff}}(k,t)
\to
\alpha+\frac{\beta}{2}
+
O\left(
\left(\frac{u_{\mathrm{cross}}}{u_{k,\infty}}\right)^2
+u_{k,\infty}^{-1}
\right).
\]

\subsection{Conditions for the existence of the asymptotic regions}
\label{app:region-existence-conditions}

We now derive conditions on the hyperparameters
\(\alpha\), \(\beta\), \(\gamma_0\), \(\sigma_w^2\), \(\sigma_a^2\), \(\lambda_0\), and \(\mu_0\) under which each region exists.

\paragraph{Region I.}
At \(t=0\), we always have \(R_k(0) = 0 < 1\), so this region always exists.

\paragraph{Region II.}
Since the upper end of this region is the \(k\)-coordinate of the intersection between \(k_{\mathrm{cross}}(t)\) and \(k_{\mathrm{learn}}(t)\), this region exists when the coordinate is much larger than \(1\). Because \(t_k^*(u)\) is monotonically increasing, the \(k\)-coordinate of the intersection is determined by
\begin{align}
\frac{0.9\gamma_0\theta_0k^{-\beta/2}}{S}
&=
\frac{\sqrt{
\sigma_w^2(2\sigma_w^2+\sigma_a^2)
}}{S}.
\end{align}
Therefore, the condition for this region to exist over a wide range within \(1 \le k \le D\) is
\begin{equation}
\left(
\frac{
0.9\gamma_0\theta_0
}{
\sqrt{
\sigma_w^2(2\sigma_w^2+\sigma_a^2)
}
}
\right)^{2/\beta}
\gg
1.
\end{equation}
Moreover, Region II requires
$u_{\mathrm{cross}}\ll u_k(t)\ll\min(1,u_{k,\infty})$.
In particular, we must have $u_{\mathrm{cross}}\ll1$,
which requires
\[
\sigma_a^2\gg\sigma_w^2.
\]
Thus, the initialization variance of the second layer must
be much larger than that of the first layer.

\paragraph{Region IIIa.}
Similarly, denoting the upper and lower boundaries of this region by \(k_u\) and \(k_l\), respectively, we have \(u_{k_u,\infty}=u_{\mathrm{cross}}\) and \(u_{k_l,\infty}=1\).
The condition for this region to exist over a broad range is
\begin{equation}
    \max\left(1,k_l\right)
    \ll
    \min\left(D,k_u\right).
    \label{eq:region-iiia-width}
\end{equation}
A necessary condition, obtained by solving \(k_l\ll k_u\), is
\begin{align}
    \frac{\gamma_0\theta_0}{S} &\ll \frac{\gamma_0\theta_0}{\sqrt{\sigma_w^2(2\sigma_w^2+\sigma_a^2)}},\\
    \therefore
    \sigma_a^2
    &\gg
    \frac{-1+\sqrt{5}}{2}\sigma_w^2.
    \label{eq:region-iiia-scale-condition}
\end{align}
In other words, the initialization scale of the second layer must be
sufficiently larger than that of the first layer for Region~IIIa to exist.
We consider sufficiently large input dimension \(D\), with all other parameters held fixed and \(\beta>0\). Since \(k_l=(\gamma_0\theta_0/S)^{2/\beta}\) is independent of \(D\), the condition \(k_l\ll D\) is automatically satisfied in this limit.

\paragraph{Region IIIb.}
The conditions for this region are
\begin{equation}
k<k_u
\qquad\text{and}\qquad
k<k_l.
\end{equation}
The required condition is therefore
\begin{equation}
\left[
\min
\left(
\frac{\gamma_0\theta_0}{S},
\frac{
\gamma_0\theta_0
}{
\sqrt{
\sigma_w^2(2\sigma_w^2+\sigma_a^2)
}
}
\right)
\right]^{2/\beta}
\gg
1.
\end{equation}

Combining the above results, the four regions coexist when
\[
\sigma_w \ll \sigma_a
\qquad\text{and}\qquad
\frac{\sigma_w^2+\sigma_a^2}{\theta_0} \ll \gamma_0 \ll \frac{D^{\beta/2}(\sigma_w^2+\sigma_a^2)}{\theta_0}.
\]
The conditions that the initialization scale of the second layer is larger than that of the first layer and that \(\gamma_0\) is large generally correspond to the feature-learning regime, in which the second layer changes significantly. Conversely, when
\[
\gamma_0 \ll \frac{\sqrt{
\sigma_w^2(2\sigma_w^2+\sigma_a^2)
}}{\theta_0},
\]
the entire range \(1\le k\le D\) belongs to Region I, and the power-law exponent of the spectrum remains nearly unchanged. In this paper, we therefore refer to this situation as the lazy regime, in contrast to the feature-learning regime.

\section{Proof that the ordering of the eigenvalues \texorpdfstring{$\rho_k(t)$}{rho k(t)} is preserved}
\label{app:ordering}

Consider the system
\begin{equation}
\left\{
\begin{aligned}
\rho_k(t)
&=
\lambda_k
\left\{
\sigma_w^2
+
\frac{S}{2}
\left[
-1
+
\sqrt{
1+\frac{4\xi_k(t)^2}{S^2}
}
\right]
\right\},
\\
\dot{\xi}_k
&=
\frac{\lambda_k}{\gamma_0}
\left(
\theta_k-\frac{\xi_k}{\gamma_0}
\right)
\sqrt{4\xi_k(t)^2+S^2},
\\
\xi_k(0)
&=
0,
\\
\lambda_k
&=
\lambda_0k^{-\alpha},
\\
\theta_k
&=
\theta_0k^{-\beta/2}.
\end{aligned}
\right.
\end{equation}
We show that if \(k>l\), then
\begin{equation}
\rho_k(t)\leq\rho_l(t)
\end{equation}
for every \(t\); that is, the ordering of the eigenvalues is preserved throughout training.

Define
\begin{equation}
\phi(\xi,\theta)
:=
\frac{1}{\gamma_0}
\left(
\theta-\frac{\xi}{\gamma_0}
\right)
\sqrt{4\xi^2+S^2},
\end{equation}
and denote by \(\xi(t;\lambda,\theta)\) the solution of the initial-value problem
\begin{equation}
\dot{\xi}
=
\lambda\phi(\xi,\theta),
\qquad
\xi(0)=0.
\end{equation}
For \(\theta\geq0\), the solution satisfies
\begin{equation}
0
\leq
\xi(t;\lambda,\theta)
\leq
\gamma_0\theta,
\end{equation}
and along this trajectory,
\begin{equation}
\phi(\xi(t;\lambda,\theta),\theta)
\geq
0.
\end{equation}
Hence, \(\xi(t;\lambda,\theta)\) is monotonically increasing in \(t\).

We next show that \(\xi(t;\lambda,\theta)\) is monotonically increasing in both \(\lambda\) and \(\theta\). Let
\begin{equation}
p
:=
\partial_{\lambda}\xi.
\end{equation}
Differentiating the initial-value problem gives
\begin{equation}
\dot{p}
=
\phi(\xi,\theta)
+
\lambda\partial_\xi\phi(\xi,\theta)p,
\qquad
p(0)=0.
\end{equation}
Its solution is
\begin{equation}
p(t)
=
\int_0^t
\phi(\xi(s),\theta)
\exp
\left[
\lambda
\int_s^t
\partial_\xi\phi(\xi(u),\theta)\,du
\right]
\,ds
\geq
0.
\end{equation}
Therefore,
\begin{equation}
\partial_{\lambda}
\xi(t;\lambda,\theta)
\geq
0.
\end{equation}

Similarly, let
\begin{equation}
q
:=
\partial_{\theta}\xi.
\end{equation}
Then
\begin{equation}
\dot{q}
=
\lambda\partial_\xi\phi(\xi,\theta)q
+
\lambda\partial_{\theta}\phi(\xi,\theta),
\qquad
q(0)=0,
\end{equation}
with
\begin{equation}
\partial_{\theta}\phi(\xi,\theta)
=
\frac{\sqrt{4\xi^2+S^2}}{\gamma_0}
\geq
0.
\end{equation}
Thus,
\begin{equation}
q(t)
=
\lambda
\int_0^t
\partial_{\theta}\phi(\xi(s),\theta)
\exp
\left[
\lambda
\int_s^t
\partial_\xi\phi(\xi(u),\theta)\,du
\right]
\,ds
\geq
0,
\end{equation}
and consequently
\begin{equation}
\partial_{\theta}
\xi(t;\lambda,\theta)
\geq
0.
\end{equation}

Finally, because
\begin{equation}
\partial_{\lambda}
\xi(t;\lambda,\theta)
\geq
0
\qquad\text{and}\qquad
\partial_{\theta}
\xi(t;\lambda,\theta)
\geq
0,
\end{equation}
and because both \(\lambda_k\) and \(\theta_k\) are ordered in descending order, \(k>l\) implies
\begin{align}
\xi_k(t)
&=
\xi(t;\lambda_k,\theta_k)
\\
&\leq
\xi(t;\lambda_l,\theta_k)
\\
&\leq
\xi(t;\lambda_l,\theta_l)
\\
&=
\xi_l(t).
\end{align}
Combining this result with the fact that \(\rho_k\) is monotonically increasing in \(\xi_k\) and \(\lambda\) gives
\begin{equation}
\rho_k(t)
\leq
\rho_l(t).
\end{equation}
Therefore, the input-dimension index \(k\) coincides with the eigenvalue rank.

\section{Numerical settings}
\label{app:numerical-settings}

For all plots, we used
$\lambda_k=k^{-1}$ and $\mu_k=\zeta(2)^{-1}k^{-1}$; equivalently,
$\alpha=\beta=\lambda_0=1$ and
$\mu_0=1/\zeta(\alpha+\beta)=1/\zeta(2)$.  Thus,
$\sum_{k=1}^{\infty}\lambda_k\mu_k=1$.  Inputs and targets were generated as
$x\sim\mathcal N(0,\Lambda)$ and $y=\Theta^\top x$, with
$\Lambda=\operatorname{diag}(\lambda_1,\ldots,\lambda_D)$ and
$\Theta=\operatorname{diag}(\sqrt{\mu_1},\ldots,\sqrt{\mu_D})$.
The initialization scale of the second layer was fixed at $\sigma_a=1.0$.

\paragraph{Figure 2.}
We trained the finite-width linear network in Eq.~(1) with
$D=P=512$, $N=8192$, and $M=4096$ fixed i.i.d.\ training samples by
full-batch gradient descent on the empirical mean-squared loss.  The remaining
parameters were
\[
(\sigma_w,\gamma_0)=\left(0.1,2.0\cdot \frac{\sigma_w^2+\sigma_a^2}{\theta_0} \right),
\]
with learning rate $\nu=N$. At each checkpoint, letting
$H_{mi}(t)=w_i(t)^\top x_m$, we plotted the nonzero eigenvalues of
$H(t)H(t)^\top/(MN)$ in decreasing order.

\paragraph{Figure 4.}
We used the same data model and teacher, and replaced the student by
\[
f_\phi(x,t)=\frac{1}{\gamma_0N}
\sum_{i=1}^{N}a_i(t)\mathrm{ReLU}\!\left(w_i(t)^\top x\right).
\]
The numerical settings were
\[
(D,P,N,M)=(512,512,8192,4096).
\]
We used $\sigma_w=0.1, \gamma_0=2.0(\sigma_w^2+\sigma_a^2)/\theta_0$ in the feature-learning regime and
$\sigma_w=1.0, \gamma_0=0.5(\sigma_w^2+\sigma_a^2)/\theta_0$ in the lazy regime.
Spectra were recorded same as Figure 2 for preactivations.

\end{document}